\pdfoutput=1

\documentclass[11pt]{article}

\usepackage[final]{acl}

\usepackage{times}
\usepackage{latexsym}

\usepackage[T1]{fontenc}

\usepackage[utf8]{inputenc}

\usepackage{microtype}

\usepackage{inconsolata}

\usepackage{amsmath,amsfonts,bm}

\def\eqref#1{equation~\ref{#1}}

\def\1{\bm{1}}

\def\vs{{\bm{s}}}

\def\mK{{\bm{K}}}

\DeclareMathAlphabet{\mathsfit}{\encodingdefault}{\sfdefault}{m}{sl}
\SetMathAlphabet{\mathsfit}{bold}{\encodingdefault}{\sfdefault}{bx}{n}

\def\sR{{\mathbb{R}}}

\newcommand{\softmax}{\mathrm{softmax}}

\DeclareMathOperator*{\argmin}{arg\,min}

\usepackage{hyperref}
\usepackage{url}

\usepackage{amsmath,amsfonts}
\usepackage{algorithmic}
\usepackage{algorithm}
\usepackage{array}
\usepackage{stfloats}
\usepackage{textcomp}
\usepackage{verbatim}
\usepackage{amsthm}
\usepackage{anyfontsize}
\usepackage{float}            
\usepackage{mathrsfs}         
\usepackage{bm}              
\usepackage{threeparttable}   
\usepackage{makecell}      

\usepackage{booktabs}
\usepackage{multirow}
\usepackage{wrapfig}
\usepackage{balance}
\usepackage{colortbl}

\usepackage{graphicx}
\usepackage{multicol} 
\usepackage{subcaption}
\usepackage{enumitem}
\usepackage{float}

\newcommand{\sysname}[0]{\textcolor{black}{EMS}}

\title{EMS: Adaptive Evict-then-Merge Strategy for Head-wise KV Cache Compression Based on Global-Local Importance}

\author{Yingxin Li$^{1}$,
        Ye Li$^{1}$, 
        Yuan Meng$^{2}$, 
        Xinzhu Ma$^{3}$,\\
        {\bf Zihan Geng$^{1}$},
        {\bf Shutao Xia$^{1}$}, 
        {\bf Zhi Wang$^{1}$} \\
$^{1}$Shenzhen International Graduate School, Tsinghua University \\ $^{2}$Department of Computer Science and Technology \\ $^{3}$Chinese University of Hong Kong \\
\small{
    \texttt{\{liyx23, liye23\}@mails.tsinghua.edu.cn,}
    \texttt{yuanmeng@tsinghua.edu.cn}, }\\
\small{
    \texttt{xinzhuma94@gmail.com,}
    \texttt{\{geng.zihan, xiast, wangzhi\}@sz.tsinghua.edu.cn
    } }
}

\begin{document}
\maketitle
\begin{abstract}
KV Cache has been widely adopted in auto-regressive LLMs to reduce the redundant computation during the inference. 
In light of growing concerns about memory overhead, efficient KV cache compression has garnered significant attention.
Most existing methods tackle this problem from two angles: identifying important tokens and developing compression strategies. 
However, these approaches often yield biased distribution of important tokens due to the influence of causal mask or positional encoding.
Moreover, they overlook the sparsity and redundancy across different heads, which leads to difficulties in preserving the most effective information at the head level. 
To this end, we propose an adaptive Evict-then-Merge Strategy (EMS) to overcome these limitations, while achieving better KV cache compression under extreme compression ratios. 
Specifically, we introduce a Global-Local score that combines accumulated attention scores from both global and local KV tokens to better identify the token importance and implement a kernel for efficiency. 
For the compression strategy, the EMS framework can account for both sparsity and redundancy across different heads. 
Additionally, we implement head-wise parallel compression through a zero-class mechanism to enhance efficiency. 
Extensive experiments demonstrate our SOTA performance even under extreme compression ratios. 
EMS improves the average score on LongBench by over 1.28 points across 4 representative LLMs under a cache budget of 256, preserves 95\% retrieval accuracy with a cache budget less than 2\% in the NIAH task, and achieves up to 10.75$\times$ throughput gains over the original FlashAttention2 implemented model.
\end{abstract}

\section{Introduction}
\label{section:intro}

\begin{figure}[t]
    \centering
    \includegraphics[scale=0.5]{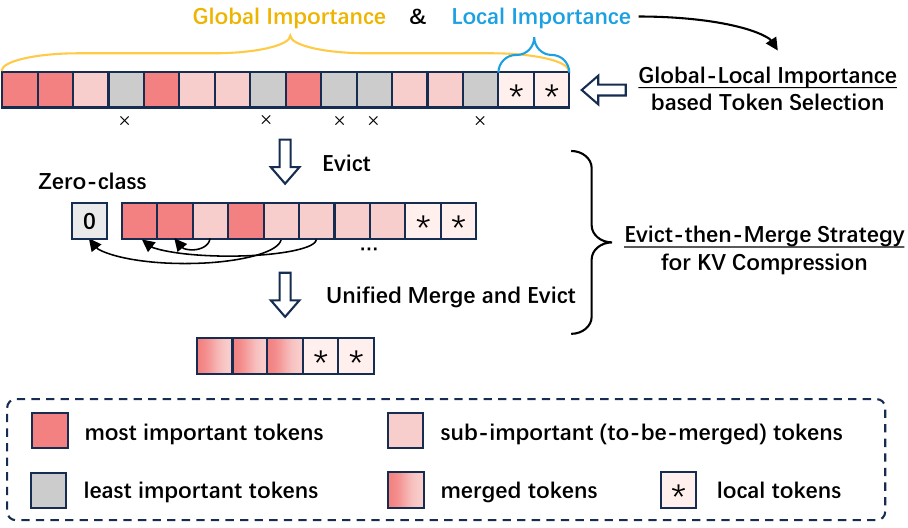}
    \caption{
    The KV cache compression workflow of~\sysname. 
    The tokens are first partitioned according to the ranking of Global-Local score, which is calculated based on attention weights of global and local tokens. 
    The least important tokens are then evicted, while the sub-important tokens are either merged into most important tokens or evicted by merging into the zero-class token. 
    }
    \label{intro_workflow}
    \vspace{-1em}
\end{figure}

Large Language Models (LLMs) \citep{devlin2018bert,GPT3,anil2023palm,llama3,mistral} have demonstrated remarkable capabilities in handling complex tasks \citep{ram2023context, chain_of_thought, roziere2023code}, driving an increasing demand for processing long sequences efficiently \citep{longlora,jin2024llm,chen2023extending}. 
To mitigate redundant computation, KV Cache is widely adopted in auto-regressive LLMs by storing previously generated key-value (KV) states, enabling faster inference. 
However, the size of the KV cache grows linearly with the length of the input sequence, which severely limits the applicability of LLMs \citep{yuan2024llm}. 
Therefore, efficiently compressing the KV cache while preserving essential information has become a critical issue \citep{Zhou2024ASO}.

Existing work mainly approaches this issue from two perspectives: ({\it i}) discerning the importance of the generated KV tokens, and ({\it ii}) devising compression strategies to preserve more information. 
For example, H2O \citep{H2O} focuses on globally important tokens, while SnapKV \citep{Snapkv} concentrates on tokens with higher relevance within a local window size. 
However, due to cumulative effects of attention weights and positional encoding, the selected tokens exhibit a biased tendency: H2O favors earlier information, while SnapKV leans towards later context.
For compression strategies, building upon lightweight token compression methods \citep{tome,A-vit,PRANCE}, existing approaches are generally categorized into evict-based and merge-based methods.
However, a unified and effective approach that combines evict and merge for extreme KV cache compression is still largely unexplored.
Based on the two limitations, we pose the following question:

\noindent \textbf{\emph{Can we select important KV tokens in a more balanced manner while retaining as much information as possible at a high compression ratio?}}


To answer this question, we decouple the compression of the KV cache into two stages: selecting important tokens and compressing the KV cache based on those selections.
Under this workflow, we propose \sysname, a head-wise {\bf E}vict-then-{\bf M}erge {\bf S}trategy based on Global-Local importance, as illustrated in Figure \ref{intro_workflow}.
Building on the token importance bias, we dynamically integrate the global- and local-aware importances and construct Global-Local score, which serves as the indicator for the subsequent compression. 
To bridge the scale gap, the score is calculated by aligning the accumulated attention scores of all KV tokens and local KV tokens.
This approach not only ensures a more balanced selection of important KV tokens, but also mitigates biases caused by attention accumulation and positional encoding. 
In addition, to enhance the efficiency of getting token importance indicator, we integrate the calculations of global and local scores into the FlashAttention2 \citep{dao2023flashattention2} kernel.

For the compression strategy, the observed difference in sparsity and redundancy across different heads suggests that applying head-wise eviction and merging can potentially achieve a higher compression ratio.
Building on this, we propose a unified Evict-then-Merge strategy at the fine-grained head level to improve the storage density.
In particular, the most irrelevant tokens will first be evicted, leaving only the tokens with higher importance scores for further compression.
A subset of the remaining tokens is then selected as class centers based on their higher importance, with less important tokens being merged into these centers.
However, as not all tokens are suitable for merging due to low redundancy, those with low similarity to important KV tokens are evicted to minimize output disturbance. 
To ensure parallel inference during head-wise merging and eviction, we introduce a zero-class center where evicted tokens are merged, treating the eviction process as a special case of merging.
Additionally, this step allows for a dynamic merge ratio for each head, ensuring more adaptive and efficient compression.
\sysname~achieves extreme compression while preserving LLM capabilities, boosting the average score on LongBench by more than 1.28 points on 4 LLMs with a 256 cache budget, and retaining 95\% retrieval accuracy using under 2\% the context length in the Needle-in-a-Haystack task.
With 4096 prompt tokens and 8192 generated tokens, \sysname~improves the throughput by 10.75× over the original FlashAttention2 implemented model.
To summarize, our contribution can be generalized as:

\setlist{topsep=0pt, partopsep=0pt, parsep=0pt, itemsep=0.1pt}

\begin{itemize}
\item [(1)] \textbf{We design a more balanced Global-Local score for important token selection.}
The score automatically integrates global and local attention, reducing bias and ensuring more balanced token selection across tasks.
\item [(2)] \textbf{We propose a sparsity- and redundancy-driven Evict-then-Merge compression strategy.} 
Leveraging the characteristic of KV tokens, it significantly enhances the information retention even under low compression ratios.
\item [(3)] \textbf{We implement an efficient head-wise parallel compression for the KV cache.}
We adapt the Global-Local score into the FlashAttention2 kernel and introduce a zero-class to unify the evicting and merging for efficiency.
\end{itemize}

\section{Related Work}
\label{section:related_work}

\subsection{Important KV Selection}
\label{section:related_work1}
Recent work has focused on the selection of important KV tokens to preserve the performance of LLMs.
StreamingLLM \citep{sink} discovers the sink mechanism and retains the first four tokens along with some local tokens, enabling streaming generation. 
Following it, ACT \citep{ACT} further demonstrates that certain KV tokens in the middle also exhibit high attention weights.
H2O \citep{H2O} selects important KV pairs based on the accumulated attention score, leading to more concentration in the former context.
SnapKV \citep{Snapkv} chooses to calculate the accumulated attention score with a local window and retains more recent tokens.
However, the valuable information is often spread across the contexts and the selection of important KV pairs should avoid former or recent preference caused by importance indicator.

\subsection{KV Compression}
\label{section:related_work2}
Inspired by various token compression methods \citep{A-vit,tome,PRANCE}, KV cache compression can also be categorized into eviction and merge strategies.
Evict-based methods aim to retain only the important KV pairs under different token importance assumption \citep{sink, ACT, H2O, Snapkv, Keyformer}.
DCP \citep{DCP} introduces a lightweight attention block in each layer to dynamically decide which KV pairs to discard. 
NACL \citep{nacl} compresses the KV cache based on some proxy tokens and random eviction.
FastGen \citep{FastGen} applies distinct compression strategies for each head through attention profilling in the prefilling stage.
There are also some works approaching this problem from the perspective of cache budget allocation. PyramidInfer \citep{yang-etal-2024-pyramidinfer}, PyramidKV \citep{cai2024pyramidkv} and CAKE \citep{qin2025cake} consider the allocation across layers, while HeadKV \citep{fu2025headkv} and DuoAttention \citep{xiao2025duoattention} focus on the allocation across heads.
These works are orthogonal to those with uniform budget strategies and can be applied to them for better compression.

Compared to evict-only methods, merge-based approaches have the potential to retain more information.
CAM \citep{CaM} analyzes the attention output error caused by token eviction and only merges the evicted value tokens into the remaining ones to reduce performance loss. 
DMC \citep{DMC} dynamically decides whether to merge the current token to the tail of the KV cache in a head-wise manner. However, compression only at the tail may not be the most effective approach. 
LESS \citep{LESS} introduces a low-rank embedding sidekick with sparse policy, which accumulates the information discarded by the eviction strategy into a fixed-size low-rank cache. 
Besides, DMC and LESS require additional training for performance gains. 
It is worth noting that KV cache merging is different from token merging, due to the paired processing of key-value tokens and the autoregressive characteristic of LLMs.
Improper merging methods can lead to significant error accumulation in the decoding stage, resulting in severe performance degradation. 
Building on these insights, we propose an efficient Evict-then-Merge compression method that effectively addresses these challenges.

\section{Hybrid Token Selection Policy based on Global-Local Score}
\label{section: glo-loc score}

\begin{figure}[t]
    \centering
  \begin{subfigure}[b]{0.9\columnwidth}
    \includegraphics[width=\columnwidth]{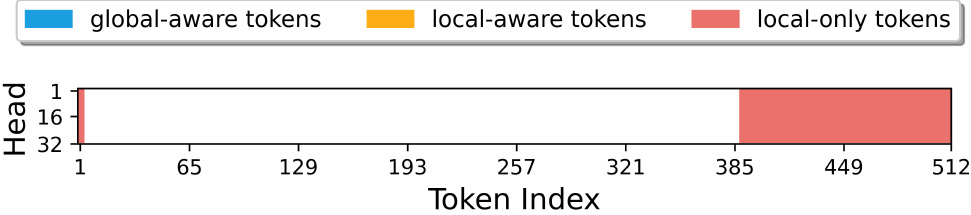}
    \vspace{-1.6em}
    \caption{\small{Local-only selection}}
    \vspace{0.3em}
    \label{Fig2a}
  \end{subfigure}
  \begin{subfigure}[b]{0.9\columnwidth}
    \includegraphics[width=\columnwidth]{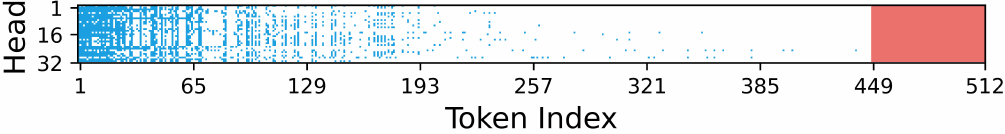}
    \vspace{-1.6em}
    \caption{\small{Global-aware selection}}
    \vspace{0.3em}
    \label{Fig2b}
  \end{subfigure}
  \begin{subfigure}[b]{0.9\columnwidth}
    \vspace{-0.3em}
    \includegraphics[width=\columnwidth]{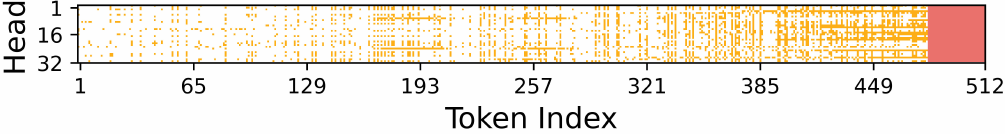}
    \vspace{-1.6em}
    \caption{\small{Local-aware selection}}
    \vspace{0.3em}
    \label{Fig2c}
  \end{subfigure}
  \begin{subfigure}[b]{0.9\columnwidth}
    \includegraphics[width=\columnwidth]{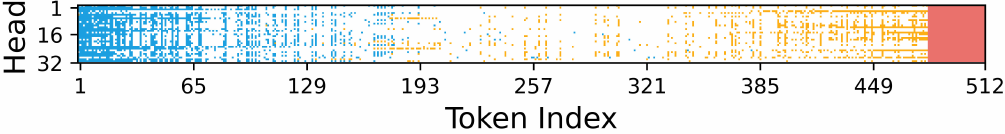}
    \vspace{-1.6em}
    \caption{\small{Global-Local based selection}}
    \vspace{0.3em}
    \label{Fig2d}
  \end{subfigure}
  \vspace{-0.5em}
    \caption{{Token selection patterns.
    The sample is taken from the \textit{gov\_report} \citep{gov_report} dataset, showing Top-128 selected tokens out of a total of 512 tokens. 
    The proposed Global-Local based selection integrates the advantages of global and local viewpoints, indicating a more balanced approach.
    }}
    \label{Fig2}
    \vspace{-1em}
\end{figure}

\begin{figure*}[t]
    \centering
    \includegraphics[scale=1]{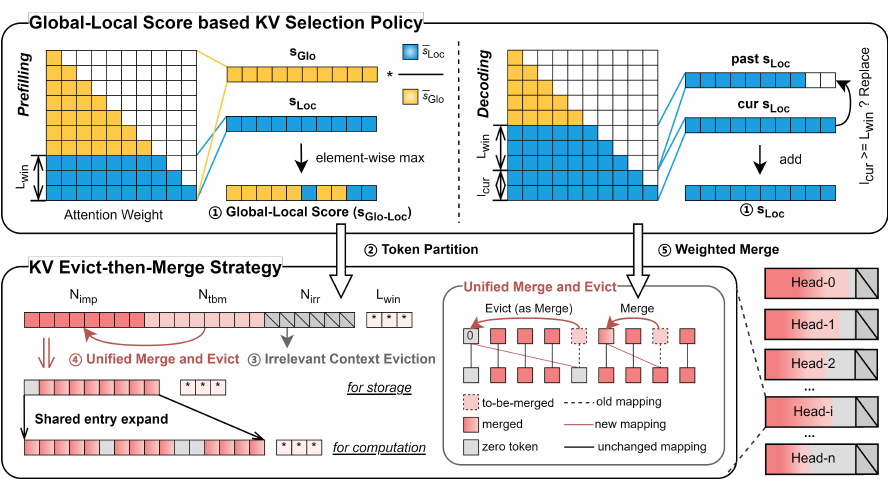}
    \caption{
    The framework of \sysname. The compression of KV cache is decoupled into two parts. For important KV selection policy, a balanced Global-Local score is designed to grasp token importance. For KV compression strategy, the Evict-then-Merge approach first removes irrelevant tokens, then applies a unified head-wise eviction and merging process.
    }
    \label{main framework}
    \vspace{-1em}
\end{figure*}

\subsection{Preliminaries}

To reduce computational redundancy, LLMs can store previously computed key and value states for future generation.
In particular, Given $n$ prompt tokens, the model first prefills the prompt information to the key and value states, and the KV cache is initialized by $\boldsymbol C^0_K = \left( \boldsymbol k_1, \boldsymbol k_2, \cdots , \boldsymbol k_n \right)$ and $\boldsymbol C^0_V = \left( \boldsymbol v_1, \boldsymbol v_2, \cdots , \boldsymbol v_n \right)$. 
During the decoding stage, the newly generated tokens will be fed into the model, the corresponding key and value states will be appended to $\boldsymbol C_{K}$ and $\boldsymbol C_{V}$.
Taking timestep $t$ as an example, the model computes $\boldsymbol q_t$, $\boldsymbol k_t$ and $\boldsymbol v_t$, and loads the previous key and value states. 
The KV cache is updated by $\boldsymbol C^t_{K}=[\boldsymbol C^{t-1}_{K}, \boldsymbol k_t]$ and $\boldsymbol C^t_{V}=[\boldsymbol C^{t-1}_{V}, \boldsymbol v_t]$, and the attention is calculated by:
\begin{equation}
\begin{gathered}
    \text{Attention}({\boldsymbol q_{t}},{\boldsymbol K_{t}},{\boldsymbol V_{t}}) =  \\ \softmax \left( {\frac{{{\boldsymbol q_{t}} \left[ {\boldsymbol C^{t-1}_K,{\boldsymbol k_{t}}} \right]^T}}{\sqrt d}} \right) \left[ {\boldsymbol C^{t-1}_V,{\boldsymbol v_{t}}} \right],
\end{gathered}
\end{equation}
where $d$ is the feature dimension. Afterwards, $\boldsymbol C_K^t$ and $\boldsymbol C_V^t$ are stored and will be loaded for the subsequent generation.
However, the memory overhead grows with the sequence, for which efficiently compressing the KV cache is critical.

\subsection{Token Importance Bias}
\label{section: glo-loc score, observation}
Existing methods for assessing the most essential KV tokens exhibit inherent biases. 
We categorized them into three types that impact compression: local-only bias, local-aware bias, and global-aware bias, as illustrated in Figure \ref{Fig2}.

Building on the attention sink \citep{sink,ACT}, \textbf{local-only methods} primarily focus on a few initial sink tokens and recent ones within a local window. 
Although they exhibit strong capabilities in local language modeling, their performance drops significantly in tasks requiring global semantic understanding. 
To select more informative KV tokens, 
\textbf{global-aware methods} \citep{H2O,Spatten} employ the global accumulated attention score to select important tokens, which can be calculated by ${{\boldsymbol s}_{Glo}} = \sum\nolimits_{i = 1}^N {{\boldsymbol A_{i,:}}}$, where $N$ is the context length, and $\boldsymbol A \in {\displaystyle \sR^{N \times N}}$ is the attention weight.
However, the cumulative effect of causal attention weight skews the importance distribution toward the earlier part, resulting in a biased selection of important KV tokens and leading to suboptimal outcomes.
In contrast, \textbf{local-aware methods} \cite{Snapkv,Scissorhands} utilize local tokens as anchors to identify and retain tokens across the entire KV cache. The accumulated attention score of local tokens is used to measure the importance, which can be formualted by ${{\boldsymbol s}_{Loc}} = \sum\nolimits_{i = N-{L_{win}}}^N {\boldsymbol A_{i,:}}$, where $L_{win}$ is the window size.
However, the positional encoding causes nearby tokens to exhibit higher correlations, leading to a bias towards retaining more recent tokens.

\subsection{Global-Local Score}
\label{section: glo-loc score, method}
To eliminate the influence of cumulative effects and positional encoding, we propose a Global-Local score to leverage both global and local information, which is shown in Figure \ref{main framework}.
Specifically, we convert ${\mathbf{s}}_{Glo}$  and ${\mathbf{s}}_{Loc}$  to similar magnitudes by mean-alignment, followed by the element-wise max to obtain the new score:
\begin{equation}
{\boldsymbol s_{Glo - Loc}} = {\max \left( {{\boldsymbol s_{Glo}} \times \frac{{\sum {{\boldsymbol s_{Loc}}} /N}}{{\sum {{\boldsymbol s_{Glo}}} /N}},{{\boldsymbol s_{Loc}}}} \right)} ,
\end{equation}
where $\max$ is an element-wise function, $N$ is the context length.

During the prefilling stage, we calculate the accumulated attention scores of $\boldsymbol A$ from global and local scope to get $\boldsymbol s_{Glo}$ and $\boldsymbol s_{Loc}$. 
These two critical vectors capture the importance distribution of the prompt and are stored for subsequent generation.
During the decoding stage, $\boldsymbol s_{Glo}$ is updated by accumulating the attention of new tokens.
The local tokens in the window will dynamically change. 
To avoid potential overhead, we set a past score $\boldsymbol s_{Loc}^{past}$ to record the attention information in the last window, and a current score $\boldsymbol s_{Loc}^{cur}$ to accumulate attention from the new queries in the current window.
Once the number of tokens in the current window reaches $L_{win}$, $\boldsymbol s_{Loc}^{cur}$ will be assigned to $\boldsymbol s_{Loc}^{past}$ and reset to zero.
The final local score $\boldsymbol s_{Loc} = \boldsymbol s_{Loc}^{past} + \boldsymbol s_{Loc}^{cur}$.
So in the decoding stage, the window size used to select local-aware tokens is actually $L_{win}\sim 2L_{win}-1$.


With the Global-Local score, the selection of important KV tokens becomes more balanced, as shown in Figure \ref{Fig2d}. In addition, we also conduct a statistic on the contribution of global and local importance to the Global-Local score in Appendix \ref{appendix: selection}.

\section{Head-wise Evict-then-Merge Strategy for KV Cache}
\label{section: evition-then-merge}

\subsection{Head-wise Sparsity and Redundancy}
\label{section: evition-then-merge, observation}

\begin{figure*}[t]
  \centering
  \vspace{-0.5em}
  \begin{subfigure}{.33\textwidth}
    \includegraphics[width=\linewidth]{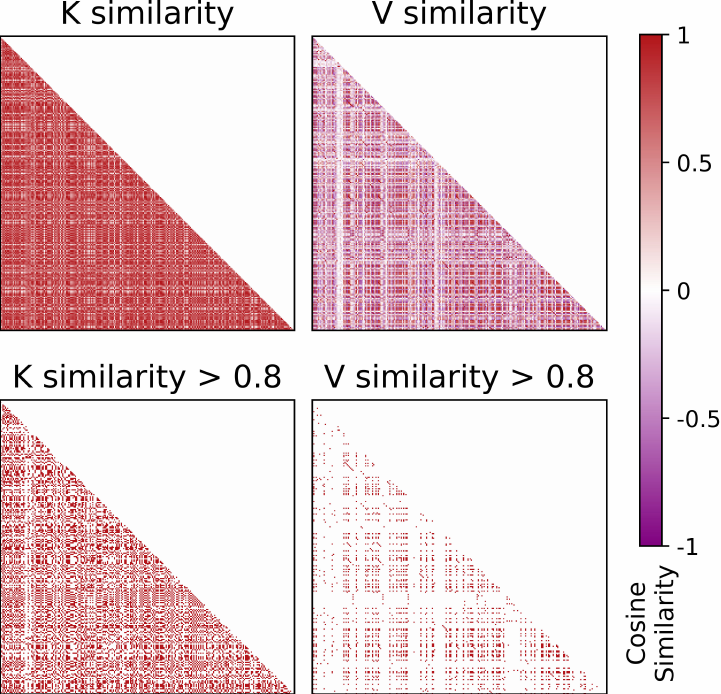}
    \caption{}
    \label{Fig3a}
  \end{subfigure}
  \hfill 
  \begin{subfigure}{.65\textwidth}
    \includegraphics[width=\linewidth]{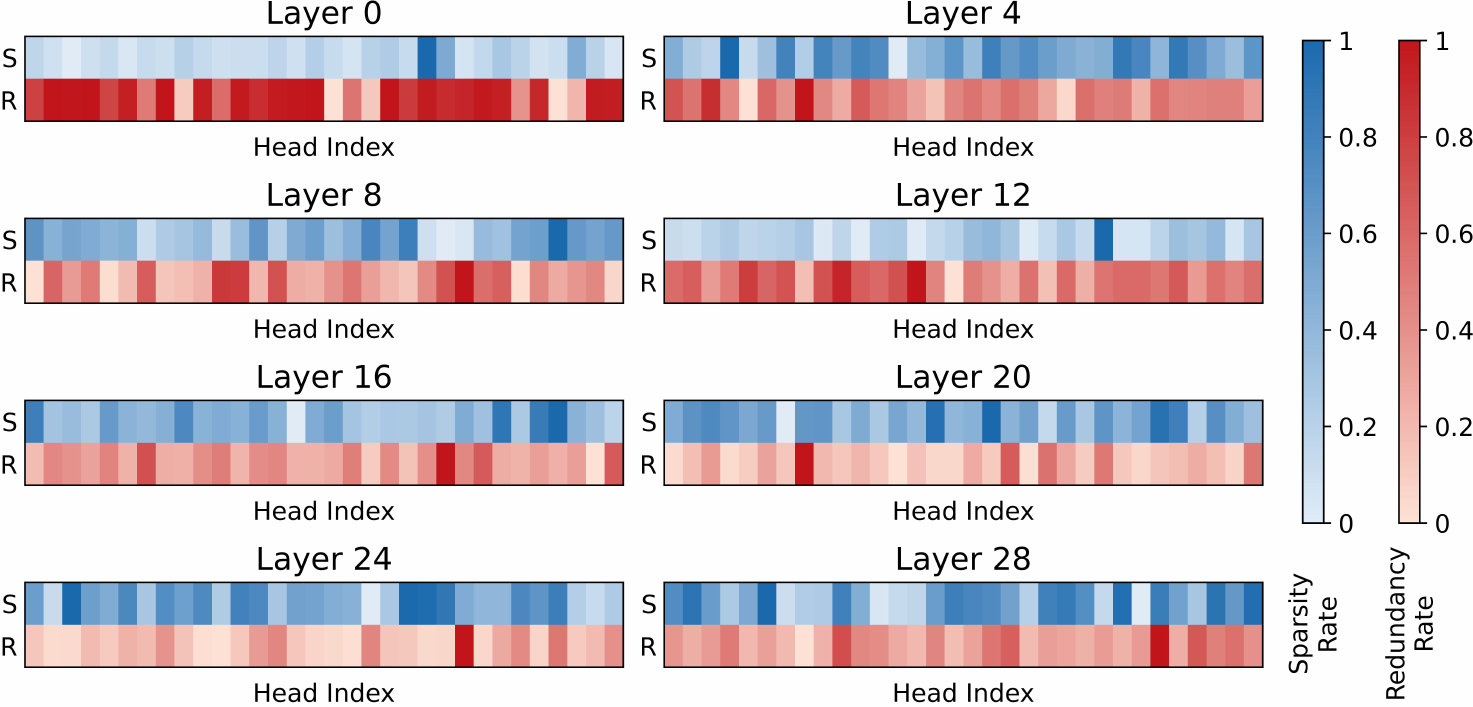}
    \caption{}
    \label{Fig3b}
  \end{subfigure}
  \vspace{-0.6em}
  \caption{
    Observations on sparsity and redundancy. The parameters $\zeta$ and $\tau$ are set to 0.95 and 0.6 here. (a) The distribution difference between key and value similarities. The top two figures depict the raw similarity, while the bottom two showcase the masked KV similarities with a threshold of 0.8. Key similarity is much more salient than value similarity. (b) The head-wise sparsity and redundancy. The blue bars represent the sparsity of each head, while the red bars denote the redundancy. Both sparsity and redundancy vary across different heads and layers.
    }
    \vspace{-0.1in}
  \label{Fig3}
\end{figure*}

Previous works \citep{H2O, Scissorhands, FastGen} have uncovered the sparsity of the KV cache. 
We further observed that the redundancy within the KV cache is also significant. 
Taking a holistic view of both sparsity and redundancy presents an opportunity to achieve higher compression ratios.

\textbf{Sparsity}. 
Only a small portion of important tokens contributes to a significant percentage of the $\boldsymbol s_{Glo-Loc}$ score. 
The sparsity rate of each attention head is determined by the minimum percentage of important tokens retained per head that can achieve over $\zeta$ of the total score:
\begin{equation}
\begin{gathered}
    {\boldsymbol p_m} = 1 - \frac{1}{N} \cdot \argmin_{N_k \in [1,N]} \left\{ {N_k\;|\;\sum\limits_{i = 1}^{{N_k}} \boldsymbol s_i  \ge \zeta} \right\}, \\ \text{where} \ \boldsymbol s = \mathrm{sort}({\boldsymbol s_{Glo - Loc}}),
\end{gathered}
\end{equation}
where $\boldsymbol p_m$ is the sparsity rate, $m$ is the head index, $\mathrm{sort}(\cdot)$ reorders the vector in descending way.

\textbf{Redundancy}. 
Cosine similarity is a reliable metric for cross position token redundancy.  
Considering that the superimposed positional encoding can attenuate the similarity, we analyze the similarity of raw KV tokens. 
As shown in Figure \ref{Fig3a}, the similarity among key tokens is significantly higher, whereas the similarity among value tokens is relatively lower.
If we consider only the key’s characteristic, the substantial merging error of value tokens can severely affect the generation.
Therefore, we jointly consider the key similarity and the value similarity, and define the redundancy $R_{i,j}$ as the product of both.
A token is considered redundant if it can find another token in the existing KV cache with redundancy exceeding the predefined threshold $\tau$. The redundancy rate $\boldsymbol r_m$ of head $m$ is given by:
\begin{equation}
\boldsymbol r_m=
\frac{1}{N}\cdot\sum_{k=1}^{N} \left [ \max ( \boldsymbol R_{k,1:k}) \ge \tau \right ].
\end{equation}
\vspace{-1em}

The head-level sparsity and redundancy rates for different heads are shown in Figure \ref{Fig3b}. 


\textbf{Head-wise Sparsity and Redundancy}. 
From an overall perspective, different heads demonstrate varying degrees of sparsity and redundancy. 
Consequently, dynamically evicting and merging tokens based on the characteristic of each head can lead to a higher compression ratio.
However, the challenge arises from the fact that all heads within a layer are stored and computed in parallel, making it difficult to manage fine-grained evict and merge decisions for each head. 
In the following section, we propose an efficient, parallelizable, and head-wise solution.

\subsection{Adaptive Evict-then-Merge Strategy}
\label{section: method}
\textbf{Token Partition}.
The Global-Local score can effectively measure the importance of tokens. 
To avoid extract fragmented information, we apply a mean pooling function to it.
Based on the ranking of $\boldsymbol s_{Glo-Loc}$, KV tokens can be divided into three sets: $N_{irr}$ irrelevant tokens $(\boldsymbol K_{irr}, \boldsymbol V_{irr})$, $N_{tbm}$ to-be-merged (TBM) tokens $(\boldsymbol K_{tbm}, \boldsymbol V_{tbm})$, and $N_{imp}$ most important tokens $(\boldsymbol K_{imp}, \boldsymbol V_{imp})$, except that $L_{win}$ local tokens $(\boldsymbol K_{loc}, \boldsymbol V_{loc})$ are always kept to preserve the local modeling capability of LLMs.
The most irrelevant tokens are first evicted to streamline the process and minimize the impact of irrelevant context \citep{irrelevant_context}.
Then, the most important tokens will serve as class centers based on the cache budget $N_{budget} = N_{imp} + L_{win}$, while $N_{tbm}$ sub-important tokens will be merged into them to preserve as much valuable information as possible. The merge magnification factor $\gamma=(N_{budget}+N_{tbm})/{N_{budget}}$ indicates the extent to which tokens are merged relative to the cache budget.

\textbf{Unified Evict-then-Merge Strategy.} 
By calculating the redundancy between TBM tokens and class center tokens, the merge destinations for TBM tokens can be identified:
\begin{equation}
    \boldsymbol d_i = \arg \max_d (\boldsymbol R_{i, d}), 
\end{equation}
where $\boldsymbol d_i$ is the merge destination for $i$-th TBM token.
To minimize output perturbation, only the tokens with redundancy $\bm{R}_{i,:d_i}$ exceeding a threshold $\tau$ will be merged, while others being evicted.
Inspired by the virtual neighbor technique \citep{li2023ddpg,li2024td3}, we introduced a zero class center to unify the merge and eviction operations, and the evicted tokens are merged into it.

Moreover, the varying sparsity and redundancy of each head results in different eviction and merge ratios, as well as varying cache sizes. 
We choose to allocate an equal budget to each head, using a limited number of entries to store class centers. 
Based on this, the merged tokens share the same entry in the KV cache, reducing the overall size of the stored KV cache. 
A position look-up-table is carefully retained in the merging process.
During the computation of each layer, a smaller cache is loaded and expanded with the help of this table, thus enlarging the context length.
More implementation details are elaborated in Appendix \ref{appendix: method details}.

To ensure that more relevant tokens dominate the merged result, we apply a weighted merge based on their attention scores.As mentioned in Section \ref{section: glo-loc score, observation}, attention scores are crucial for identifying token importance. 
Key tokens are the decisive factor in calculating the attention score.
Therefore, we preserve the norm of key tokens, a scalar for each token, and only merge the normalized keys to maintain the accuracy of the attention weights. Value tokens are merged without being normalized for the sake of simplicity.


\section{Experiment}
\label{section: experiment}

\subsection{Settings}
\label{section: experiment, settings}
In this paper, we use LLaMA-2-7B-Chat \citep{llama2}, LLaMA-3-8B-Instruct \citep{llama3}, LongChat-7B-v1.5-32k \citep{longchat} and Mistral-7B-Instruct-v0.2 \citep{mistral} models. 
For comparison, we benchmark our method against StreamingLLM, CAM, H2O, and SnapKV. 
All these methods compress the KV cache during both the prefilling and decoding stages, except for SnapKV, which compresses only the prompt after the prefilling stage.
All experiments can be conducted on a single NVIDIA A100 GPU with 40GB of memory, except for the fully cached model.
For common parameters, both \sysname~and SnapKV use $L_{win}=32$, $\text{kernel\_size}=7$ for perplexity and LongBench, and $L_{win}=16$, $\text{kernel\_size}=7$ for Needle-in-a-Haystack task. 

\subsection{Performance Evaluation on LongBench}
\label{section: experiment, longbench}

\begin{table*}[!t]
  \centering
  \caption{{Performance evaluation on LongBench across four LLMs.} All methods are tested under cache budget 256, except for SnapKV, which increases this budget at the decoding stage. The best results are highlighted with {\bf bold}.}
  \vspace{-0.5em}
  \resizebox{\textwidth}{!}{
    \begin{tabular} 
    {c@{\hspace{1ex}}|l@{\hspace{0.05ex}}c@{\hspace{0.05ex}}c@{\hspace{0.05ex}}c@{\hspace{0.05ex}}c@{\hspace{0.05ex}}c@{\hspace{0.05ex}}c@{\hspace{0.05ex}}c@{\hspace{0.05ex}}c@{\hspace{0.05ex}}c@{\hspace{0.05ex}}c@{\hspace{0.05ex}}c@{\hspace{0.05ex}}c@{\hspace{0.05ex}}c@{\hspace{0.5ex}}c@{\hspace{0.6ex}}c@{\hspace{0.6ex}}c@{\hspace{0.6ex}}c}
    \specialrule{1.5pt}{0pt}{2pt}
    
      & \multirow{4}{*}{\parbox{2cm}{\centering \textbf{Method}}}  & \multicolumn{3}{c}{\textbf{Multi-Document QA}} & \multicolumn{3}{c}{\textbf{Single-Document QA}}& \multicolumn{3}{c}{\textbf{Summarization}}& \multicolumn{3}{c}{\textbf{Few-shot Learning}}& \multicolumn{2}{c}{\textbf{Synthetic}} & \multicolumn{2}{c}{\textbf{Code}} & \multirow{4}{*}{\textbf{Avg.}} \\
    \cmidrule(lr){3-5}\cmidrule(lr){6-8}\cmidrule(lr){9-11}\cmidrule(lr){12-14}\cmidrule(lr){15-16}\cmidrule(lr){17-18}
    & & \rotatebox[origin=c]{30}{HotpotQA} & \rotatebox[origin=c]{30}{2WikiMQA} & \rotatebox[origin=c]{30}{Musique} & \rotatebox[origin=c]{30}{MF-en} & \rotatebox[origin=c]{30}{NrtvQA} & \rotatebox[origin=c]{30}{Qasper} & \rotatebox[origin=c]{30}{GovReport} & \rotatebox[origin=c]{30}{QMSum} & \rotatebox[origin=c]{30}{MultiNews} & \rotatebox[origin=c]{30}{TriviaQA} & \rotatebox[origin=c]{30}{SAMSum} & \rotatebox[origin=c]{30}{TREC} & \rotatebox[origin=c]{30}{PRe} & \rotatebox[origin=c]{30}{PCount} & \rotatebox[origin=c]{30}{Lcc} & \rotatebox[origin=c]{30}{RB-P} & \\

    \midrule
    \multirow{7}[4]{*}{\rotatebox{90}{\parbox{2.5cm}{\centering \textbf{Llama-2-7B-Chat}}}} & Full Cache & 30.20  & 27.37  & 11.54  & 34.33  & 19.68  & 19.40  & 24.60  & 20.81  & 26.19  & 84.61  & 41.00  & 63.50  & 8.00  & 4.50  & 61.46  & 55.14  & 33.27  \\ \cmidrule{2-19}          
    & StreamingLLM & 24.41  & 26.90  & 7.70  & 17.44  & 13.69  & 15.07  & 16.73  & 19.21  & 18.25  & 81.67  & 35.03  & 39.50  & 6.00  & 4.00  & 55.22  & 51.00  & 26.99  \\
    & H2O   & 29.92  & 25.07  & 10.46  & 23.05  & 17.04  & \textbf{18.45 } & 19.96  & 20.07  & \textbf{24.12 } & 82.66  & 38.26  & 59.50  & 3.50  & 4.00  & 56.01  & 51.64  & 30.23  \\
    & CAM   & 24.68  & 26.80  & 7.57  & 16.79  & 14.11  & 15.98  & 16.57  & 19.09  & 18.35  & 81.09  & 35.17  & 39.50  & 5.50  & 4.00  & 55.34  & 51.15  & 26.98  \\
    & SnapKV & 28.96  & 26.20  & 11.14 & 28.41  & 15.72  & 17.10  & 17.01  & 20.30  & 21.53  & 84.32  & 38.62  & 57.00  & 8.50  & 4.00  & 57.26  & 53.03  & 30.57  \\
    & \sysname(w.o. pos) & \textbf{31.39 } & 27.46  & 10.48  & \textbf{30.45 } & \textbf{18.52 } & 17.87  & \textbf{22.18 } & \textbf{20.82 } & 22.30  & 84.86  & \textbf{39.34 } & \textbf{61.00 } & 6.00  & \textbf{5.00 } & 57.86  & \textbf{54.10 } & \textbf{31.85 } \\
    & \sysname(w. pos) & 30.93  & \textbf{27.58 } & \textbf{11.37 } & 29.85  & 17.49  & 17.17  & 18.59  & 20.57  & 22.05  & \textbf{85.60 } & 38.79  & 60.50  & \textbf{9.50 } & 4.50  & \textbf{58.90 } & 53.55  & 31.68  \\
    \midrule
    
    \multirow{7}[4]{*}{\rotatebox{90}{\parbox{2.5cm}{\centering \textbf{Llama-3-8B-Instruct}}}} 
    & Full Cache & 44.93  & 37.92  & 24.10  & 41.89  & 22.84  & 39.26  & 28.69  & 23.57  & 26.58  & 90.31  & 42.67  & 74.50  & 67.00  & 6.48  & 57.13  & 51.34  & 42.45  \\
\cmidrule{2-2}          
    \cmidrule{2-19}  
    & StreamingLLM & 37.64  & 25.25  & 16.94  & 22.61  & 16.58  & 18.73  & 18.42  & 20.15  & 19.25  & 78.43  & 39.29  & 53.00  & 65.08  & 7.25  & 56.83  & 53.72  & 34.32  \\
    & H2O   & 43.74  & 34.06  & 20.65  & 27.71  & 20.34  & 26.67  & 21.27  & 20.58  & 23.98  & 88.52  & 38.50  & 58.50  & 66.50  & \textbf{7.50 } & 57.14  & 52.11  & 37.99  \\
    & CAM   & 37.64  & 25.19  & 16.94  & 22.64  & 16.58  & 18.58  & 18.49  & 20.10  & 19.24  & 78.43  & 39.34  & 53.00  & 65.08  & 7.25  & 56.83  & 53.56  & 34.31  \\
    & SnapKV & 41.96  & 31.82  & 20.06  & 34.91  & 20.52  & 26.59  & 19.94  & 21.72  & 21.94  & \textbf{90.47 } & 39.55  & 50.50  & \textbf{67.00 } & 6.84  & 58.23  & 53.81  & 37.87  \\
    & \sysname(w.o. pos) & \textbf{45.51 } & \textbf{36.86 } & \textbf{22.80 } & \textbf{37.31 } & \textbf{21.33 } & \textbf{27.74 } & \textbf{22.11 } & \textbf{22.28 } & \textbf{24.05 } & 90.44  & \textbf{40.26 } & \textbf{64.50 } & 66.67  & 7.22  & 57.01  & 50.37  & \textbf{39.78 } \\
    & \sysname(w. pos) & 44.09  & 32.70  & 21.77  & 35.67  & 20.84  & 26.87  & 20.71  & 21.89  & 23.89  & 89.75  & 39.96  & 61.00  & \textbf{67.00 } & 7.24  & \textbf{60.51 } & \textbf{56.46 } & 39.40  \\
    \midrule
    
    \multirow{7}[4]{*}{\rotatebox{90}{\parbox{2.5cm}{\centering \textbf{LongChat-7B-v1.5-32k}}}} 
     & Full Cache & 31.26  & 22.52  & 12.16  & 43.92  & 18.33  & 28.81  & 31.29  & 22.57  & 26.32  & 82.41  & 40.06  & 66.00  & 31.50  & 0.00  & 53.02  & 55.28  & 35.34  \\
    \cmidrule{2-19}          
    & StreamingLLM & 13.06  & 9.62  & 1.97  & 4.48  & 4.02  & 8.64  & 1.00  & 3.18  & 0.92  & 4.72  & 6.96  & 20.50  & 2.96  & 0.50  & 2.16  & 5.93  & 5.66  \\
    & H2O   & 11.33  & 10.37  & 1.20  & 7.13  & 4.04  & 8.59  & 2.42  & 9.52  & 2.92  & 23.26  & 7.11  & 27.50  & 2.45  & 0.15  & 9.30  & 8.24  & 8.47  \\
    & CAM & 13.40 &	9.17 &	1.82 &	5.07 &	4.08 &	8.62 &	1.01 &	3.24 &	0.79 &	4.23 &	7.12 &	21.50 &	3.55 &	0.50 &	2.05 &	5.98 & 5.76 \\
    & SnapKV & 19.60  & 13.45  & 8.39  & 15.06  & 9.07  & 9.19  & 3.76  & 13.15  & 3.42  & 57.43  & 18.26  & 30.50  & 1.00  & 0.44  & 15.60  & 20.08  & 14.90  \\
    & \sysname(w.o. pos) & \textbf{32.07 } & 23.75  & \textbf{12.34 } & \textbf{39.93 } & 16.07  & 22.96  & \textbf{23.89 } & \textbf{21.22 } & \textbf{23.49 } & \textbf{79.58 } & \textbf{36.89 } & 61.50  & 11.50  & 0.00  & \textbf{55.08 } & 50.78  & 31.94  \\
    & \sysname(w. pos) & 31.51  & \textbf{23.75 } & \textbf{12.34 } & 37.16  & \textbf{16.39 } & \textbf{23.47 } & 20.33  & 21.18  & 22.83  & 78.70  & 36.76  & \textbf{61.50 } & \textbf{27.75 } & \textbf{0.60 } & 54.02  & \textbf{52.39 } & \textbf{32.54 } \\
    \midrule
    
    \multirow{7}[4]{*}{\rotatebox{90}{\parbox{2.5cm}{\centering \textbf{Mistral-7B-Instruct-v0.2}}}} 
    & Full Cache & 36.42  & 21.77  & 19.13  & 47.12  & 21.02  & 29.62  & 32.57  & 24.02  & 27.09  & 86.23  & 42.99  & 71.00  & 89.33  & 3.07  & 54.00  & 51.87  & 41.08  \\
    \cmidrule{2-19}          
    & StreamingLLM & 21.60  & 13.24  & 10.25  & 26.35  & 13.72  & 11.58  & 17.98  & 19.73  & 18.92  & 80.67  & 40.26  & 50.50  & 24.80  & \textbf{3.82 } & 50.57  & 44.04  & 28.00  \\
    & H2O   & 23.31  & 14.07  & 10.32  & 33.42  & 14.29  & 16.75  & 23.12  & 21.09  & 23.73  & 83.22  & 38.74  & 63.00  & 31.50  & 3.44  & 49.17  & 44.66  & 30.86  \\
    & CAM   & 21.15 &	13.25 &	10.25 &	26.41 &	13.72 &	11.6 &	17.88 &	19.72 &	18.94 &	80.7 &	40.19 &	50.5 &	25.05 &	3.82 &	50.6 &	44.04 &	27.99 \\
    & SnapKV & 26.28  & 14.58  & 12.02  & 40.82  & 16.88  & 18.87  & 21.46  & 21.60  & 22.05  & 84.82  & \textbf{40.79 } & 51.00  & \textbf{70.08 } & 2.72  & 51.01  & 46.75  & 33.86  \\
    & \sysname(w.o. pos) & 25.83  & 15.36  & 12.38  & 40.99  & \textbf{17.44 } & 18.84  & \textbf{23.13 } & 22.21  & \textbf{24.07 } & 84.94  & 40.28  & \textbf{64.00 } & 60.77  & \textbf{4.13 } & \textbf{51.87 } & \textbf{47.59 } & 34.61  \\
    & \sysname(w. pos) & \textbf{26.49 } & \textbf{15.91 } & \textbf{12.57 } & \textbf{41.51 } & 17.28  & \textbf{18.90 } & 22.63  & \textbf{22.50 } & 23.82  & \textbf{85.17 } & 40.44  & 63.50  & 69.40  & 3.74  & 51.49  & 46.88  & \textbf{35.14 } \\
    \specialrule{1.5pt}{1pt}{0pt}
    \end{tabular}%
  }
  \label{table1}%
  \vspace{-0.5em}
\end{table*}

LongBench \citep{longbench} is a comprehensive benchmark designed to assess long context modeling abilities across 6 categories of tasks. 
To demonstrate the effectiveness of \sysname~under extreme KV cache compression, we enforce a strict compression budget of 256, with $\tau$ and $\gamma$ set to approximately 0.6 and 4. 
To adapt the calculation of global attention score on a single GPU for long-context models, we compute scores by retaining half of the tokens from both the start and the end, following observations from “Lost in the Middle” \citep{lost_in_the_middle}, where LLMs tend to have a better grasp of the information at the beginning or the end.

The results are shown in Table \ref{table1}. We present two versions of our method: one without position information to merge more aggressively and capture global information, and one with position information for tasks requiring precise token localization. 
As shown in the table, the baseline methods show varying degrees of performance across different tasks, and the optimal compression strategy differs between models. 
In contrast, our method integrates the strengths of the baselines, achieving the state-of-the-art performance on nearly all tasks, with improvements of 1.28, 1.79, 17.64, and 1.28 across the four different LLMs. 
Notably, on LongChat-7B-v1.5-32k, while the performance of other compression methods breaks down, our method consistently maintains high performance, demonstrating its robustness under extreme compression ratios and long-context conditions.
The results under more budgets can be found in Table \ref{cache budgets}. Across various budgets, EMS outperforms baselines on the four LLMs.

\subsection{Needle-in-a-Haystack}
\label{section: experiment, needle}

\begin{table}[t]
  \centering
    \caption{{The performance of Needle-in-a-Haystack across three budgets.}}
    \vspace{-0.5em}
    \scalebox{0.9}{
    \begin{tabular}{l c c c}
     \midrule[1.2pt]
    $\textbf{N}_{\textbf{budget}}$ &  \textbf{128} & \textbf{256} & \textbf{512}  \\
    \midrule
    H2O      & 0.312 & 0.335 & 0.384 \\
    SnapKV   & 0.802 & 0.893 & 0.956 \\
    \sysname & \textbf{0.818} & \textbf{0.896} & \textbf{0.959} \\
  \midrule[1.2pt]
    \end{tabular}
   }
    \label{tab:needle}
    \vspace{-1em}
\end{table}

Needle-in-a-Haystack \citep{needle} is a challenging task to assess the model’s ability to retrieve specific information from a large volume of data.  
We test our method using Mistral-7B-Instruct-v0.2 with 10 depths, 40 lengths and maximum token limit 32k. The merge threshold $\tau$ and merge magnification factor $\gamma$ are set to 0.55 and 4. 
As shown in Table \ref{tab:needle}, \sysname~delivers the best retrieval performance across all three compression budgets. 
It retains 95.9\% retrieval ability of the fully cached model, even surpassing SnapKV, which is specifically designed for retrieval tasks. The visualization of retrieval accuracy is shown in Appendix \ref{appendix: needle}.


\subsection{Efficiency Analysis}
\label{section: experiment, efficiency}

\begin{table*}[t]
  \centering
    \captionof{table}{Comparisons of the end-to-end latency (s) on long context settings. With the constant budget, \sysname~supports larger batch sizes and generalize to longer context without OOM errors.}
    \vspace{-0.5em}
    \scalebox{0.9}{
    \begin{tabular}{c c c c c c c c}
    \midrule[1.2pt]
        \multirow{2}{*}{$\textbf{L}_{\textbf{prompt}}$ + $\textbf{L}_{\textbf{gen}}$} & \multirow{2}{*}{\textbf{Method}} & \multicolumn{5}{c}{\textbf{Batch Size}} & \multirow{2}{*}{\makecell{\textbf{Max Throughput}\\[1ex]\textbf{(tokens/s)}}} \\
        \cmidrule(lr){3-7}
        ~ & ~ & \textbf{1} & \textbf{2} & \textbf{4} & \textbf{8} & \textbf{16} & ~ \\
        \midrule
        \multirow{2}{*}{\textbf{128 + 4096}} & Full Cache & 453.7 & 558.3 & OOM & OOM & OOM & 14.67 \\ 
        ~ & \sysname & 421.7 & 442.2 & 476.6 & 706.7 & 1299.2 & 50.44 \textcolor{blue}{(3.44$\times$)} \\
        \midrule
        \multirow{2}{*}{\textbf{128 + 6144}} & Full Cache & 657.7 & 1157.6 & OOM & OOM & OOM & 10.62 \\
        ~ & \sysname & 638.7 & 676.9 & 706.0 & 1080.2 & 2010.8 & 48.89 \textcolor{blue}{(4.61$\times$)} \\
        \midrule
        \multirow{2}{*}{\textbf{4096 + 128}} & Full Cache & 17.3 & 30.8 & OOM & OOM & OOM & 8.31 \\
        ~ & \sysname & 14.1 & 15.5 & 17.5 & 28.5 & 54.9 & 37.30 \textcolor{blue}{(4.48$\times$)} \\
        \midrule
        \multirow{2}{*}{\textbf{4096 + 4096}} & Full Cache & 751.5 & 1381.1 & OOM & OOM & OOM & 5.93 \\
        ~ & \sysname & 439.1 & 458.9 & 478.9 & 761.7 & 1432.8 & 45.74 \textcolor{blue}{(7.71$\times$)} \\
        \midrule
        \multirow{2}{*}{\textbf{4096 + 8192}} & Full Cache & 1918.6 & OOM & OOM & OOM & OOM & 4.27 \\
        ~ & \sysname & 980.0 & 910.7 & 959.8 & 1518.2 & 2855.7 & 45.90 \textcolor{blue}{(10.75$\times$)} \\
        \midrule[1.2pt]
    \end{tabular}
   }
    \label{tab: time}
    \vspace{-1em}
\end{table*}

\textbf{Time analysis}.
To assess the efficiency of \sysname, we measured the end-to-end latency on two RTX 4090 GPUs with the cache budget 256. 
As shown in Table \ref{tab: time}, by compressing the KV cache, \sysname~supports larger batch sizes, whereas the fully cached model encounters out-of-memory (OOM) errors with batch sizes of 2 or 4.
Moreover, the efficiency gains of our method become more pronounced with larger batch sizes and longer context length, leading to increased throughput. 
For example, with 4096 prompt tokens and 8192 generated tokens, our method achieves a 10.75$\times$ improvement in throughput compared to the fully cached model.

\textbf{Memory analysis}. 
\sysname~compresses the storage overhead for the full KV cache $dN_{full}$ to a constant $dN_{budget}$ per head.
The extra static memory overhead is required for $\displaystyle \vs_{Glo}$, $\displaystyle \vs_{Loc}^{past}$, $\displaystyle \vs_{Loc}^{cur}$, $\|\displaystyle \mK\|$ and the mapping look-up-table. Each of these is represented as a scaler value per token.  
During the computation time of certain layer, only $N_{budget}$ key-value states are loaded and expanded to $\gamma N_{budget}$ as the runtime KV cache.
This not only enables longer contexts, but also decreases the overall memory overhead memory movement, supporting a higher throughput.

\subsection{Ablation Study}
\label{section: experiment, ablation}
We ablate the effectiveness of the Global-Local score and the Evict-then-Merge strategy, and evaluate the performance scalability across different cache budgets $N_{budget}$ using the LongBench average score on Llama-2-7B-Chat.



\begin{table}
    \centering
    \caption{{KV cache compression using different token importance scores and compression strategies.}}
    \vspace{-0.5em}
    \scalebox{0.9}{
    \begin{tabular}{c c}
     \midrule[1.2pt]
    \textbf{Score + Compression} & \textbf{Avg.}   \\
    \midrule
    $\displaystyle \vs_{Glo}$ + Evict-then-Merge & 28.81 \\
    $\displaystyle \vs_{Loc}$ + Evict-then-Merge & 31.06 \\
    $\displaystyle \vs_{Glo-Loc}$ + Evict-only & 30.82 \\
    $\displaystyle \vs_{Glo-Loc}$ + Evict-then-Merge & 31.34 \\
  \midrule[1.2pt]
    \end{tabular}
   }
    \label{tab: ablation_method}
    \vspace{-0.5em}
\end{table}

\begin{table}
  \centering
    \captionof{table}{{The impact of different cache budgets for each head. The performance scales with the budgets.}}
    \vspace{-0.5em}
    \scalebox{0.9}{
    \begin{tabular}{lcccccccccccccc}
     \midrule[1.2pt]
    $\textbf{N}_{\textbf{budget}}$ &  \textbf{128} & \textbf{256} & \textbf{512} & \textbf{768} & \textbf{1024} \\
    \midrule
    \textbf{Avg.} &  29.41 & 31.34 & 32.26 & 32.71 & 33.00 \\
  \midrule[1.2pt]
    \end{tabular}
   }
    \label{tab: cache budget}
    \vspace{-1em}
\end{table}

\textbf{Ablation on Token Selection and Compression Strategy.} 
Table \ref{tab: ablation_method} illustrated the effectiveness of our method. 
By incorporating the Evict-then-Merge strategy, the average score improves 0.61 points compared to the evict-only approach.
Furthermore, integrating both global and local scores increases the average score by 0.37 and 2.62 points, respectively, compared to using either the global or local score alone. 
These results highlight the effectiveness of our method in token selection and compression strategy.

\textbf{Cache Budgets.} 
In Table \ref{tab: cache budget}, we explore the impact of different cache budgets. 
The results indicate that even under extreme compression settings, performance remains relatively stable. 
Additionally, performance scales with cache size, approaching that of the fully cached model as the budget increases. 
When $N_{budget}=1024$, \sysname~ performs a difference of 0.27 from the full cache model. More results on different LLMs can be seen in Table \ref{cache budgets}.

Besides, we conduct the ablation and analysis on merge threshold $\tau$ and merge magnification factor $\gamma$, which affect merge-evict ratio and merge size, which are shown in Appendix \ref{appendix: parameter ablation}.

\section{Conclusion}
\label{section: conclusion}
In this paper, we propose \sysname, an input-aware, head-wise efficient KV cache management framework. 
By leveraging the proposed Global-Local attention score, \sysname~addresses the biased distribution of important KV tokens caused by accumulated attention scores and positional encoding, leading to a more balanced selection of tokens.
We further design a unified Evict-then-Merge strategy based on the redundancy and sparsity intrinsic of KV tokens across different heads. 
In particular, we implement a zero-class mechanism to enable parallel computation for head-wise operation.
Extensive experiments on language modeling perplexity, LongBench, and Needle-in-a-Haystack tasks demonstrate the SOTA performance, validating the effectiveness of \sysname. 

\section{Limitations}
\label{Limitation}

While \sysname~demonstrates strong performance, certain limitations remain. The Global-Local score provides a simple yet effective way to balance the former and recent biases in token retention. We align their mean values to achieve this balance, but this solution is derived from empirical observations rather than a theoretical foundation. A more rigorous exploration of unbiased strategy warrants further investigation. Additionally, \sysname~can be integrated with other cache budget allocation techniques, which may further enhance efficiency and performance. Conducting additional experiments would validate the effectiveness and robustness of ~\sysname. 

\bibliography{acl_latex}

\appendix
\section{Method Details and Discussion}
\label{appendix: method details}

\begin{figure}[t]
  \centering
  \begin{subfigure}{.49\textwidth}
    \includegraphics[width=\linewidth]{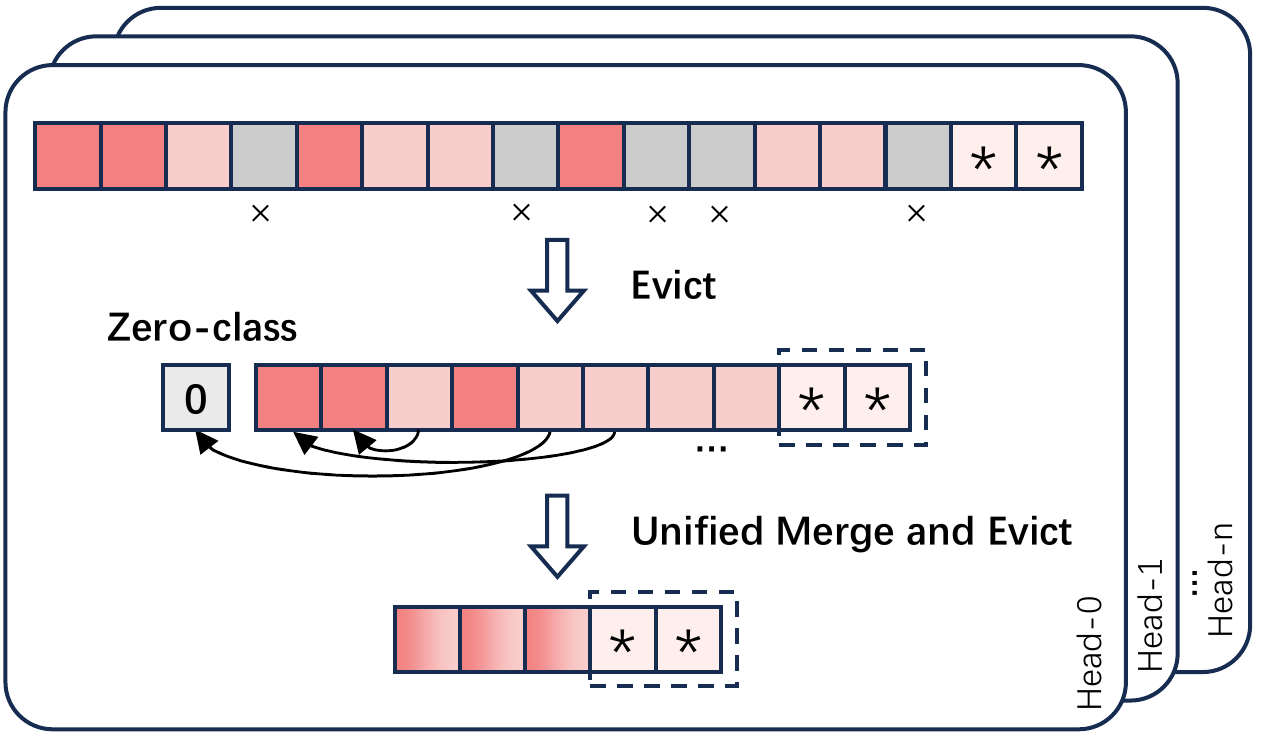}
    \caption{Two levels of eviction.}
    \label{appendix: two-level eviction}
  \end{subfigure}
  \hfill 
  \begin{subfigure}{.49\textwidth}
    \includegraphics[width=\linewidth]{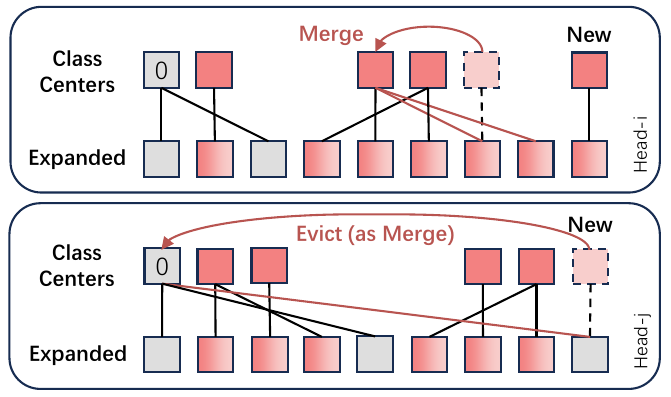}
    \caption{Unified merge and evict at decoding stage.}
    \label{appendix: merge-evict decoding}
  \end{subfigure}
  \caption{
    Evict-then-Merge details. (a) Two levels of eviction. The first level of eviction is evicting the same number of irrelevant tokens. The second level of merge is merging the tokens with low similarity to zero-class token. Different heads have different eviction at the second level. (b) Unified merge and evict at decoding stage. Different heads in the same layer have different merge or evict decisions, which are unified as merge operation.
    }
  \label{fig: method details}
  \hspace{0.3em}
\end{figure}

\textbf{Two-Levels of Parallelism.} We regarded \sysname~ as a parallel solution for two reasons. On the one hand, as illustrated in Section \ref{section: evition-then-merge, observation}, the sparsity and redundancy provide the potential for more extreme compression. However, the head-wise characteristic can lead to different cache budgets for different heads. For code implementation, the shape of KV states is \texttt{(batch\_size, num\_heads, kv\_len, head\_dim)}, which means that all heads are concatenated as a tensor and computed in parallel. Luckily, there are two properties and they exhibit complementarity to some extent. For example, we can evict more for the heads with higher sparsity and less redundancy (dark blue and light red bars), and merge more for the heads with higher redundancy and less redundancy (light blue and dark red bars). So we take this chance to allocate the same budget for each head. On the other hand, merge and evict are two different operations. Different heads may have different operations and different merge-evict ratios. To unify the merge and evict process, we introduce a zero-class, thus the eviction can be treated as merging, and we only do merging at the second stage of Evict-then-Merge strategy.

\textbf{Two Levels of Eviction.} Under our Evict-then-Merge strategy, there are two levels of eviction consideration, which is shown in Figure \ref{appendix: two-level eviction}. Under the long context scenarios, not all tokens are necessary for language modeling and can lead to significant memory overhead. Hence, the less relevant tokens will first be evicted and leaves $N_{imp}+N_{tbm}+N_{loc}$ tokens for subsequent merging, which is same for each heads. At the merge stage, considering that some TBM tokens cannot find a suitable merge destination for low similarity, evicting them can avoid disturbing the class centers. So these tokens are merged to zero-class, which is equivalent to eviction.

\textbf{Evict-then-Merge Details.} There are some implementation differences in the compression of prefilling and decoding stage. 
Specifically, at prefilling stage, numerous tokens are filled at once, which might far exceed the size of the cache budget. Therefore, it is necessary to partition them and determine the TBM tokens and class-center tokens. And this stage often requires to merge multiple tokens for each head. The merged $N_{imp}$ tokens serve as the class centers, where $N_{tbm}+N_{imp}$ tokens will share the $N_{imp}$ entries.
At decoding stage, we focus more on the update of class centers and the mapping relation, which is shown in Figure \ref{appendix: merge-evict decoding}. To achieve dynamic class center, the merge operation is processed at the class center level, which means that merging will change the mapping of the TBM tokens and class centers. After each decoding in the auto-regressive generation, a new token will exceed local range and is regarded as the class center. One least important class center token is selected as TBM token. For each head, there are two kinds of operations for this TBM token. Different heads may have different decisions for merge or evict operation, and they are unified as merge. Since we keep the cache budget constant, the number of expanded tokens is also fixed. So we need to evict one element from the look-up-table and fill the new mapping. 

\textbf{Shared Entry Expansion.} The motivation of expansion is that we hope to conduct clustering on the KV tokens based on cosine similarity. In this way, we can use a small number of class center tokens to represent more tokens. Therefore, those similar tokens share the same KV entry and need to be expanded during computation. For the implementation of expansion, we keep a position look-up-table and $N_{imp}$ tokens are expanded to $\gamma N_{imp}$ tokens at computation time of the certain layer. In this way, more context tokens participate in the computation, and only additional overhead is only brought to the current computed layer.

\textbf{Global-Local Score Efficiency Discussion.} FlashAttention2 has become a standard technique for long sequence inference. It achieves more efficient attention computation through tiling, yet doesn't return the attention weight. 
Since the calculation of the global score necessitates obtaining all the attention weights, this poses a challenge.
To resolve this contradiction, we modified the FlashAttention2 kernel and reuse the long-sum-exponent (\texttt{lse}) meta-information. By recalculating the query and key within the kernel and modifying the loop order, we efficiently computed the global and local accumulated attention scores.
We have conducted tests to show its efficiency. When \texttt{(batch\_size,num\_head,seq\_len,head\_dim) = (1,32,4096,128)}, using the modified kernel to calculate the global score on a single RTX 4090 GPU achieves a speed-up of more than 13 times compared to the original three-step calculation process (query and key multiplication, softmax operation, and reduce sum).
Thus, the dilemma regarding the global score can be solved. 
We will release the kernel along with source code once the paper is accepted. 

\section{Ablation Study}
\label{appendix: parameter ablation}

\begin{table*}[ht]
    \centering
    \caption{{Effects of merge threshold $\tau$ on LongBench. The TBM tokens with a redundancy score above $\tau$ are merged, while those below are evicted. A lower threshold results in more tokens being merged.}}
    \scalebox{0.82}{
    \begin{tabular}{lcccccccccccccc}
     \midrule[1.2pt]
    $\tau$ &  0 & 0.1 & 0.2 & 0.3 & 0.4 & 0.5 & 0.55 & 0.6 & 0.65 & 0.7 & 0.75 & 0.8 & 0.9 & 1  \\
    \midrule
    Avg. &  29.68 & 29.59 & 29.84 & 30.25 & 30.69 & 30.93 & 30.88 & \textbf{31.34} & 31.28 & 31.25 & 31.27 & 30.49 & 31.13 & 30.82 \\
  \midrule[1.2pt]
    \end{tabular}
   }
    \label{tab: merge thrd}
\end{table*}

\begin{table*}[ht!]
  \centering
    \captionof{table}{{Ablation on $\gamma$, which affects the merge size. 'Inf' means there is no eviction of irrelevant contexts and '1' means only apply eviction using $\displaystyle \vs_{Glo-Loc}$.}}
    \scalebox{0.82}{
    \begin{tabular}{lcccccccccccccc}
     \midrule[1.2pt]
    $\gamma$ &  1 & 2 & 3 & 4 & 5 & 6 & 7 & 8 & Inf \\
    \midrule
    Avg. &  30.05 & 30.88 & 31.00 & \textbf{31.34} & \textbf{31.34} & 31.25 & 31.04 & 30.92 & 31.18 \\
  \midrule[1.2pt]
  \vspace{-0.2in}
    \end{tabular}
    \label{tab: merge size}
   }
\end{table*}


\textbf{Merge Threshold.} 
The merge threshold plays a crucial role in determining whether a token will be merged into the zero class (i.e., evicted) or common class centers. A smaller threshold results in more tokens being evicted, while a larger threshold leads to more aggressive merging. 
Table \ref{tab: merge thrd} presents the results of varying the merge threshold, indicating that a moderate threshold $\tau = 0.6$, yields best performance on balance.

Two extreme cases are setting threshold to 0 or 1, meaning all merging or all evicting. If $\tau=0$, all TBM tokens are merged without considering that some tokens are not suitable to merge, leading to 1.66 performance drop. On contrary, if $\tau=1$, all TBM tokens are evicted, resulting in a performance degradation of 0.52. Therefore, both over merging and over evicting yields sub-optimal performance, manifesting the effectiveness of joint merge and evict.

\textbf{Merge Size.} 
In our evict-then-merge strategy, the number of tokens retained for class center and merging (i.e $\gamma N_{budget}$) after the initial eviction stage is also a key factor, which is indicated by merge magnification factor $\gamma$. As shown in Table \ref{tab: merge size}, we ablate the effect of $\gamma$ and find that merging approximately $3\sim 4$ times of the class center size achieve the optimal performance. 

Two extreme cases are setting $\gamma$ to 1 or Inf, which means only evict irrelevant tokens and no eviction of irrelevant tokens. When $\gamma=1$, we only apply the first level of eviction and skip the unified merge and evict operation.
On the contrast, when $\gamma=\text{Inf}$, we skip the eviction of irrelevant tokens and only implement the unified merge and evict at a longer token length. The results demonstrate that both over-merging and under-merging can negatively impact the performance. Removing some irrelevant tokens at an appropriate degree not only improves performance, but also reduces the complexity of managing the mapping relation.

\begin{table}[t]
  \centering
    \captionof{table}{ The comparison of using different similarities. The performance metric is the average score on LongBench using Llama2.}
    \scalebox{0.82}{
    \begin{tabular}{lccc}
     \midrule[1.2pt]
    $Similarity$ &  Key & Value & Key-Value \\
    \midrule
    Avg. &  31.18 & 31.26 & \textbf{31.34} \\
  \midrule[1.2pt]
  \vspace{-0.3in}
    \end{tabular}
    \label{tab: sim}
   }
\end{table}

\textbf{Key Similarity, Value Similarity and Key-Value Similarity.} We conduct experiments to evaluate the performance of merging based on key, value and key-value similarity. The results arc displayed in Table \ref{tab: sim}, which manifest the superiority of considering both key similarity and value similarity.

\textbf{Cache Budgets.} More results on the LongBench performance of different models using more budgets and their comparisons to baselines are shown in Table \ref{cache budgets}. $\gamma$ and $\tau$ are set to 4 and 0.6 as ablated. EMS outperforms other methods on the 4 LLMs across different cache budgets. 

\begin{table}[t]
  \centering
  \caption{Impact of varying cache budget across four LLMs. \sysname~outperforms other methods on the 128, 256, 512, 768 and 1024 cache budgets.}
  \scalebox{0.75}{
    \begin{tabular} {clccccc}
    
    \specialrule{1.5pt}{0pt}{2pt}

    & \textbf{Method} & \textbf{128} & \textbf{256} & \textbf{512} & \textbf{768} & \textbf{1024} \\

    \midrule
    \multirow{5}[3]{*}{\rotatebox{90}{\parbox{2cm}{\centering \textbf{Llama-2}}}} 
    & StreamingLLM & 24.39 & 26.99 & 28.92 & 29.46 & 29.77  \\       
    & H2O & 19.00 & 30.23 & 31.14 & 31.54 & 31.91  \\
    & SnapKV & 27.78 & 30.57 & 31.64 & 32.05 & 32.27  \\
    & \sysname(w.o. pos) & \textbf{29.41} & \textbf{31.34} & \textbf{32.26} & \textbf{32.71} & \textbf{33.00} \\
    & \sysname(w. pos) & 29.12 & 30.93 & 31.99 & 32.53 & 32.68 \\
    \midrule
    
    \multirow{5}[3]{*}{\rotatebox{90}{\parbox{2cm}{\centering \textbf{Llama-3}}}} 
    & StreamingLLM & 32.42 & 34.32 & 36.18 & 37.27 & 38.80  \\       
    & H2O & 36.02 & 37.99 & 39.55 & 40.20 & 40.62  \\
    & SnapKV & 35.54 & 37.87 & 39.78 & 40.57 & 40.95 \\
    & \sysname(w.o. pos) & \textbf{36.45} & \textbf{38.94} & \textbf{40.45} & \textbf{40.74} & \textbf{41.12} \\
    & \sysname(w. pos) & 36.14 & 38.62 & 40.12 & 40.73 & 41.10 \\
    \midrule
    
    \multirow{5}[3]{*}{\rotatebox{90}{\parbox{2cm}{\centering \textbf{LongChat}}}} 
     & StreamingLLM & 5.77 & 5.66 & 5.45 & 8.51 & 13.98  \\       
    & H2O & 5.71 & 8.47 & 17.71 & 26.25 & 31.19  \\
    & SnapKV & 6.04 & 14.90 & 32.08 & 33.27 & 33.63 \\
    & \sysname(w.o. pos) & \textbf{25.65} & \textbf{31.25} & 33.00 & 33.05 & 33.36 \\
    & \sysname(w. pos) & 18.85 & 31.10 & \textbf{33.32} & \textbf{34.18} & \textbf{34.01} \\
    \midrule
    
    \multirow{5}[3]{*}{\rotatebox{90}{\parbox{2cm}{\centering \textbf{Mistral}}}} 
    & StreamingLLM & 26.97 & 28.00 & 29.90 & 30.74 & 31.37  \\       
    & H2O & 29.89 & 30.86 & 32.02 & 33.38 & 34.24  \\
    & SnapKV & 31.15 & 33.86 & 36.16 & 37.24 & 37.91 \\
    & \sysname(w.o. pos) & 31.16 & 33.93 & 35.37 & 36.30 & 37.14 \\
    & \sysname(w. pos) & \textbf{31.44} & \textbf{34.66} & \textbf{36.59} & \textbf{37.64} & \textbf{38.35} \\
    \specialrule{1.5pt}{1pt}{0pt}
    \end{tabular}%
  }
  \label{cache budgets}%
\end{table}

\section{Language Modeling Perplexity}
\label{appendix: ppl}

We evaluate the perplexity on the PG19 \citep{pg19} dataset using the LLaMA-2-7B model. 
The experiment is conducted under three cache budgets: 2\%, 5\%, and 10\% of the pretraining length (i.e. 4096 for LLaMA-2-7B). The testing sequence length is extended 10 times of the original size, allowing us to thoroughly assess the model’s continuous modeling capability for long contexts. The merge threshold $\tau$ and merge magnification factor $\gamma$ are set to 0.6 and 4.
No protection is applied, allowing more aggressive merging. 

\begin{figure}
  \centering
   \includegraphics[scale=0.37]{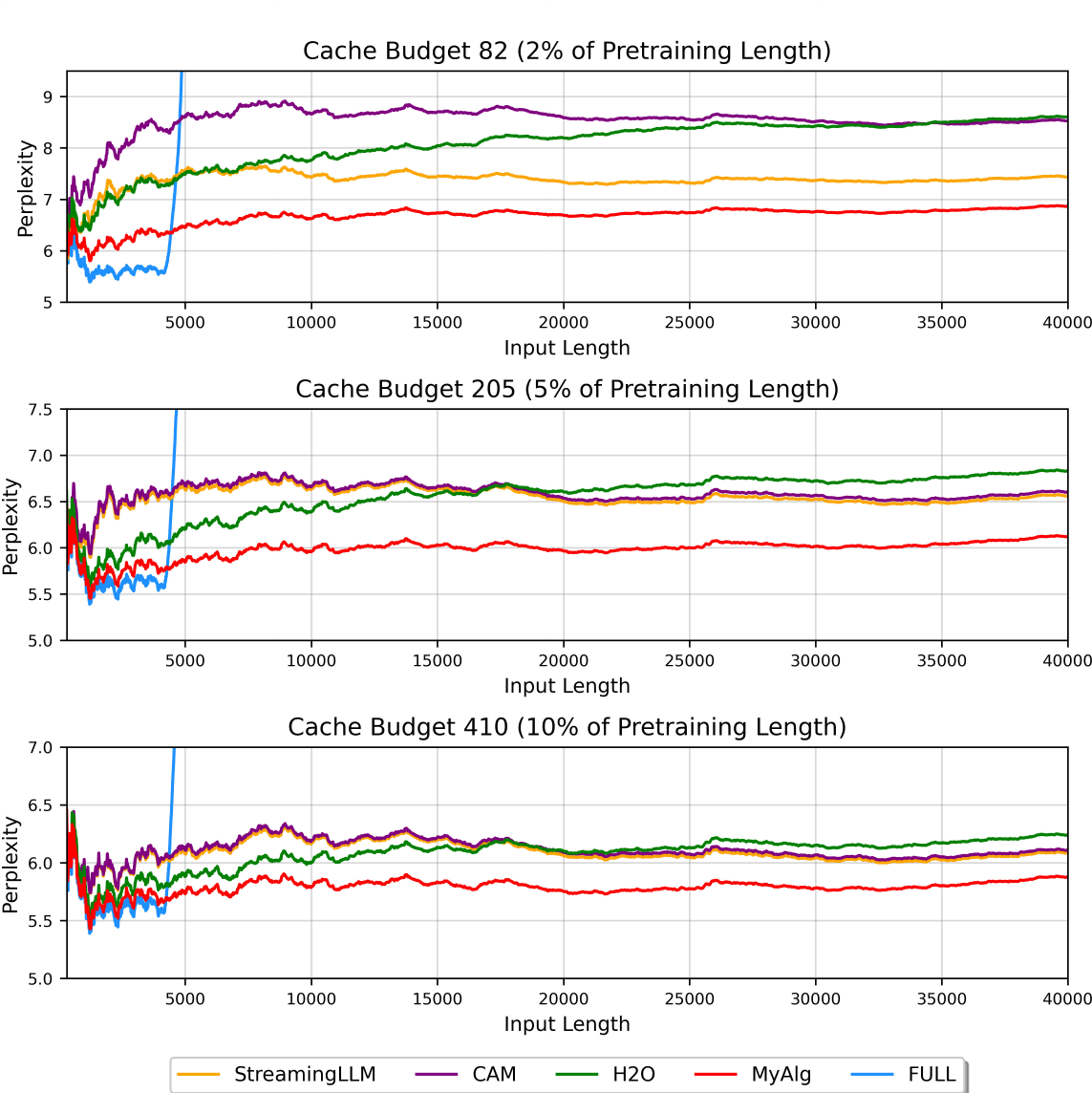}
    \caption{
    Perplexity across different cache budgets. Lower perplexity indicates better model performance. SnapKV is not listed for its lack of compression in decoding stage.
    } 
    \label{Fig4}
\end{figure}

As shown in Figure \ref{Fig4}, \sysname~consistently maintains the lowest perplexity across all budget settings. This demonstrates its robustness in managing limited cache resources while preserving language modeling accuracy. The fully cached baseline encounters significant degradation in performance once the sequence length exceeds the pretraining limit.

\section{Needle-in-a-Haystack Visualization}
\label{appendix: needle}

\begin{figure}[H]
    \centering
    \begin{subfigure}[b]{0.49\textwidth}
        \includegraphics[width=\textwidth]{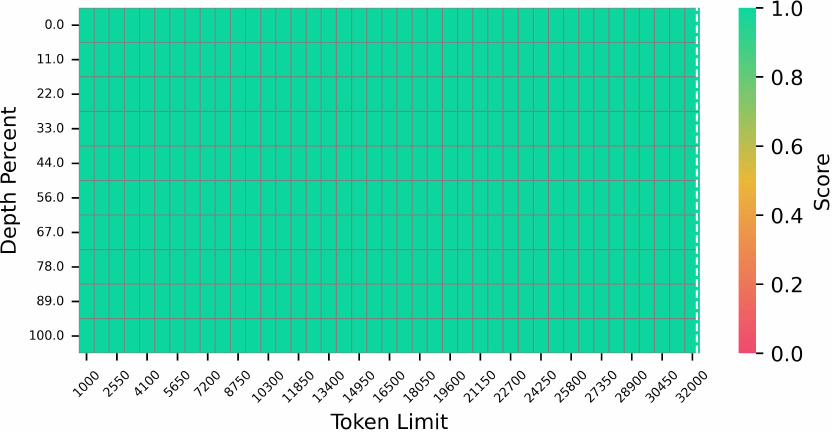}
        \caption{Full cache}
        \label{needle_fig:full}
    \end{subfigure}
    \begin{subfigure}[b]{0.49\textwidth}
        \includegraphics[width=\textwidth]{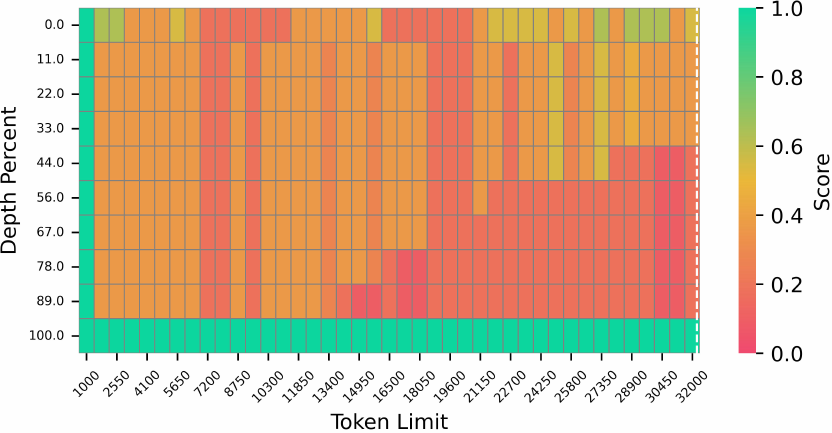}
        \caption{H2O}
        \label{needle_fig:h2o}
    \end{subfigure}

    \begin{subfigure}[b]{0.49\textwidth}
        \includegraphics[width=\textwidth]{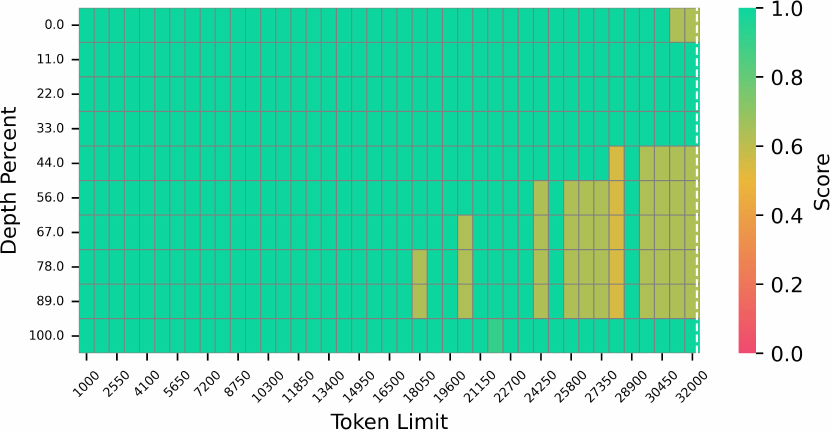}
        \caption{SnapKV}
        \label{needle_fig:snapkv}
    \end{subfigure}
    \begin{subfigure}[b]{0.49\textwidth}
        \includegraphics[width=\textwidth]{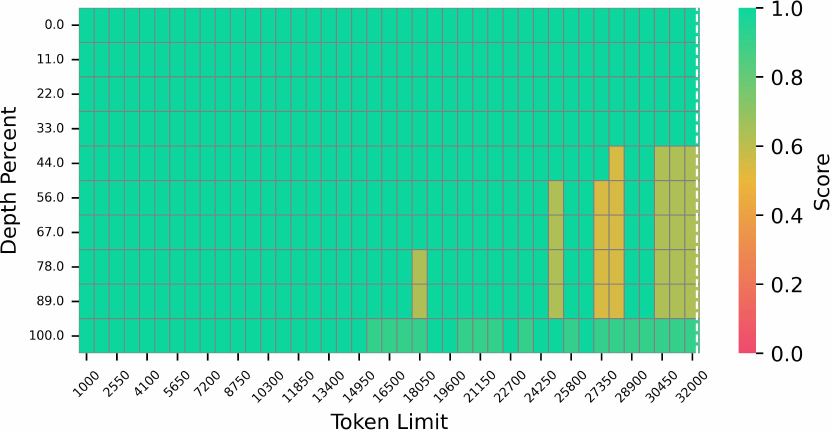}
        \caption{\sysname}
        \label{needle_fig:ours}
    \end{subfigure}
    \caption{Pressure testing results on Mistral-7B-Instruct-v1.5 with full cache and three compression methods. The maximum test length is 32k, which is almost max context length for popular LLMs. For compression methods, the cache budget for each head is 512, which is only 1.6\% of the maximum testing length.
    }
    \label{fig:needle}
\end{figure}

Needle-in-a-Haystack is particularly challenging as it requires precise retrieval from extensive context, simulating real-world scenarios where relevant information is buried among irrelevant data. Table \ref{tab:needle} has shown the numerical results, here we visualize the comparisons of baselines on retrieval accuracy across the depths and token limits under budget 512, which is shown in Figure \ref{fig:needle}. For \sysname, the
merge threshold $\tau$ and merge magnification factor $\gamma$ are set to 0.6 and 4.

We can see that H2O almost collapses on retrieval task, with the accuracy only 38.4\%. 
SnapKV, a method designed for retrieval tasks, achieves 95.6\% accuracy, while \sysname~can achieve 96.8\% retrieval ability of the fully cache model. 

\section{Token Selection across Layers}
\label{appendix: selection}

In Figure \ref{fig:selection_pattern}, we visualize more token selection patterns of different methods across layers using Llama2 and Llama3. The experiment is conducted on multi-document QA dataset \textit{hotpotqa} \citep{yang2018hotpotqa} and summarization dataset \textit{gov\_report} \citep{gov_report}. The heads of Llama3 are expanded 4 times due to the use of GQA \citep{ainslie2023gqa}.

\begin{figure}[H]
    \centering
    \begin{subfigure}{\columnwidth}
        \includegraphics[width=\columnwidth]{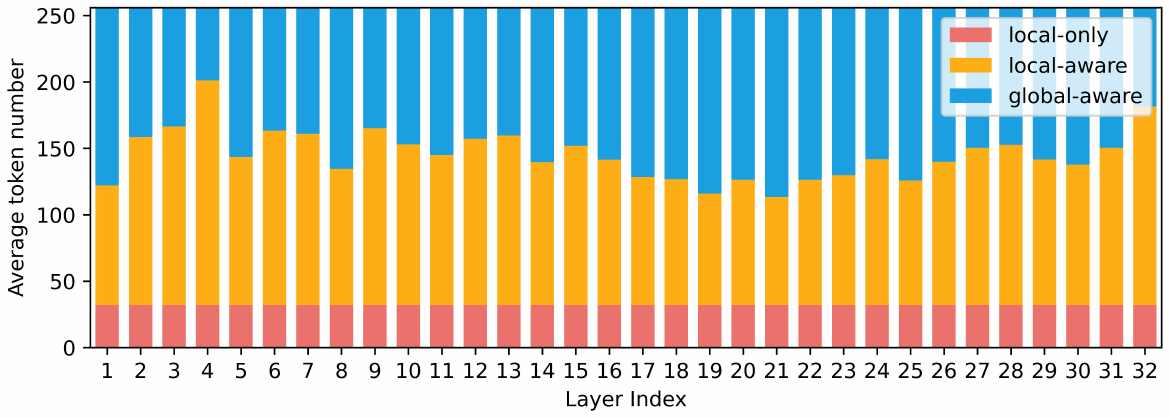}
        \caption{Select Top-256 out of 4096 tokens.}
        \label{sel_256}
    \end{subfigure}


    \begin{subfigure}{\columnwidth}
        \includegraphics[width=\columnwidth]{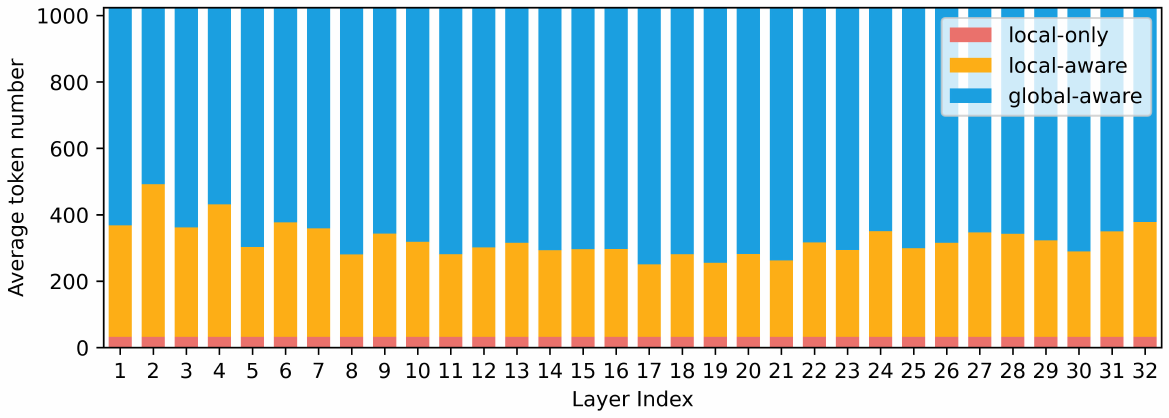}
        \caption{Select Top-1024 out of 4096 tokens.}
        \label{sel_1024}
    \end{subfigure}
    
    \caption{The distribution of selected tokens. The sample is taken from the \textit{gov\_report} dataset. When selecting fewer tokens, the Global-Local based selection method tends to evenly choose between global-aware and local-aware tokens. And when selecting more tokens, it leans toward selecting more global-aware tokens.}
    \label{fig:selection_distribution}
\end{figure}

\begin{figure*}
    \centering
    \begin{subfigure}[b]{0.47\textwidth}
        \includegraphics[width=\textwidth]{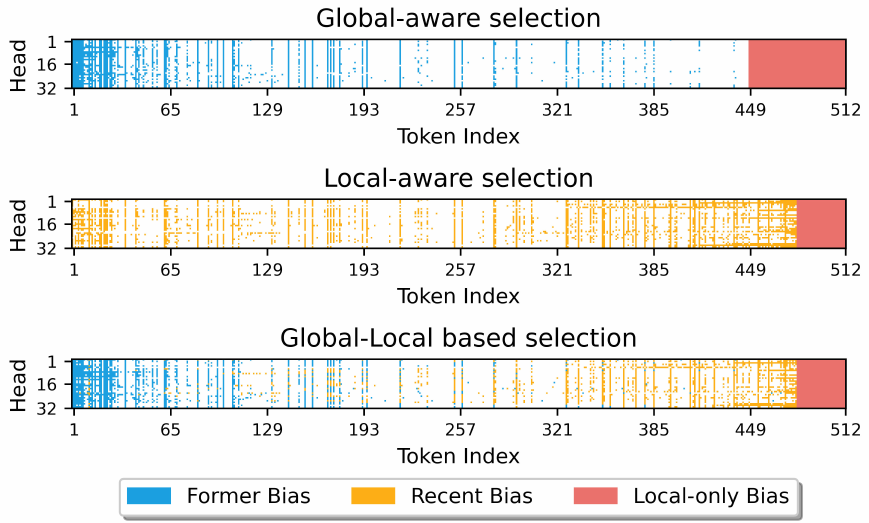}
        \caption{Layer 1 on \textit{gov\_report} using Llama2.}
        \label{selection_gov_llama2_1}
    \end{subfigure}
    \begin{subfigure}[b]{0.47\textwidth}
        \includegraphics[width=\textwidth]{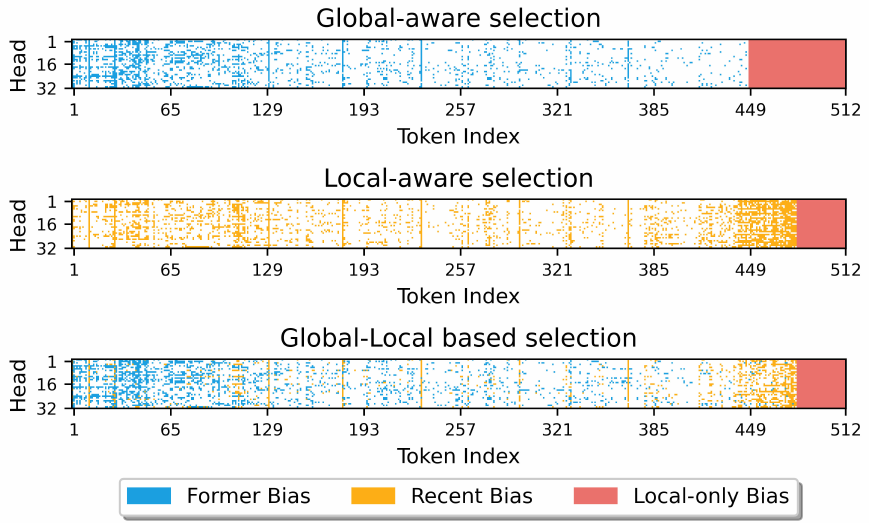}
        \caption{Layer 25 on \textit{gov\_report} using Llama2.}
        \label{selection_gov_llama2_2}
    \end{subfigure}

    \vspace{0.7em}

    \begin{subfigure}[b]{0.47\textwidth}
        \includegraphics[width=\textwidth]{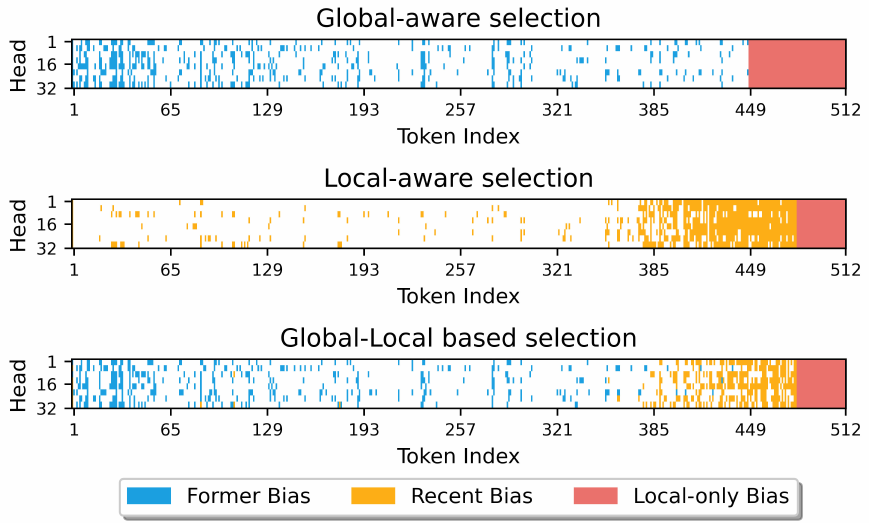}
        \caption{Layer 8 on \textit{gov\_report} using Llama3.}
        \label{selection_gov_llama3_1}
    \end{subfigure}
    \begin{subfigure}[b]{0.47\textwidth}
        \includegraphics[width=\textwidth]{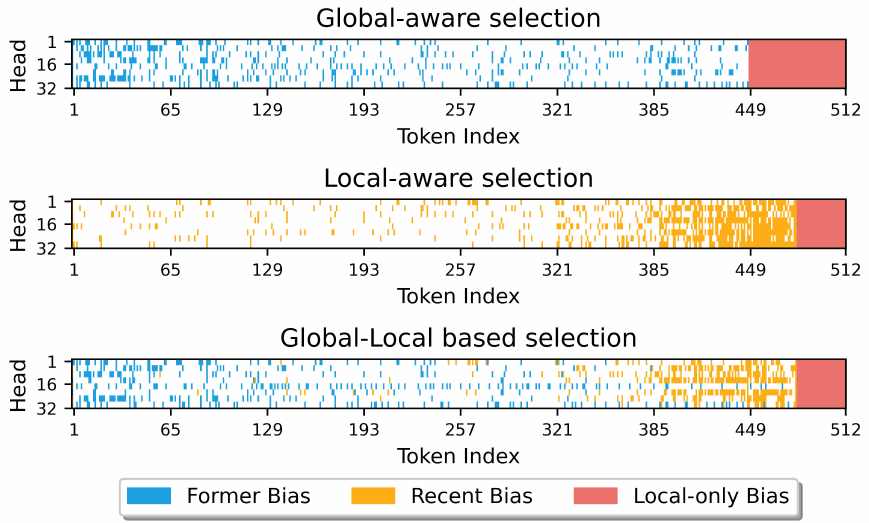}
        \caption{Layer 23 on \textit{gov\_report} using Llama3.}
        \label{selection_gov_llama3_2}
    \end{subfigure}

    \vspace{0.7em}
    
    \begin{subfigure}[b]{0.47\textwidth}
        \includegraphics[width=\textwidth]{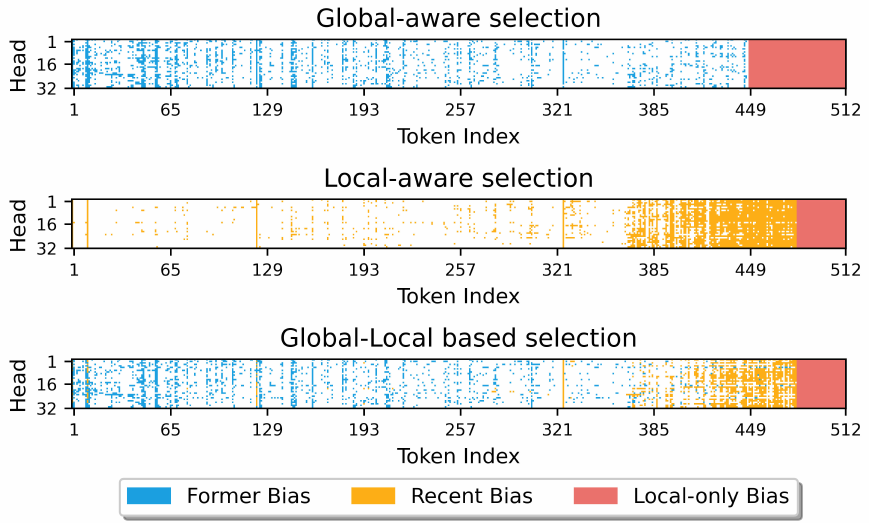}
        \caption{Layer 10 on \textit{hotpotqa} using Llama2.}
        \label{selection_hotpotqa_llama2_1}
    \end{subfigure}
    \begin{subfigure}[b]{0.47\textwidth}
        \includegraphics[width=\textwidth]{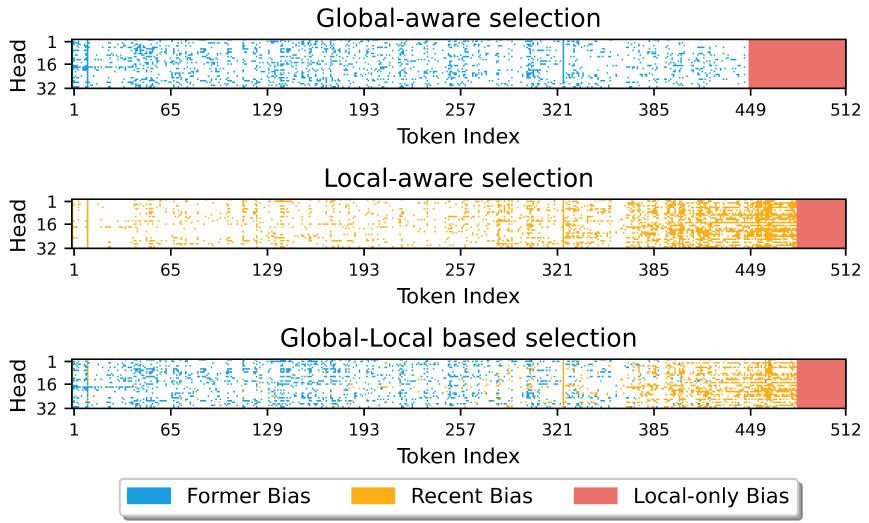}
        \caption{Layer 29 on \textit{hotpotqa} using Llama2.}
        \label{selection_hotpotqa_llama2_2}
    \end{subfigure}
    
    \vspace{0.7em}

    \begin{subfigure}[b]{0.47\textwidth}
        \includegraphics[width=\textwidth]{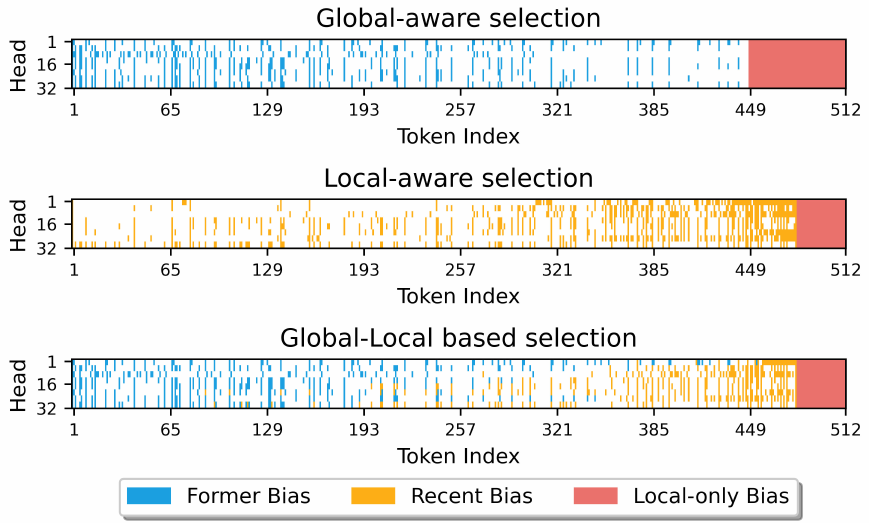}
        \caption{Layer 0 on \textit{hotpotqa} using Llama3.}
        \label{selection_hotpotqa_llama3_1}
    \end{subfigure}
    \begin{subfigure}[b]{0.47\textwidth}
        \includegraphics[width=\textwidth]{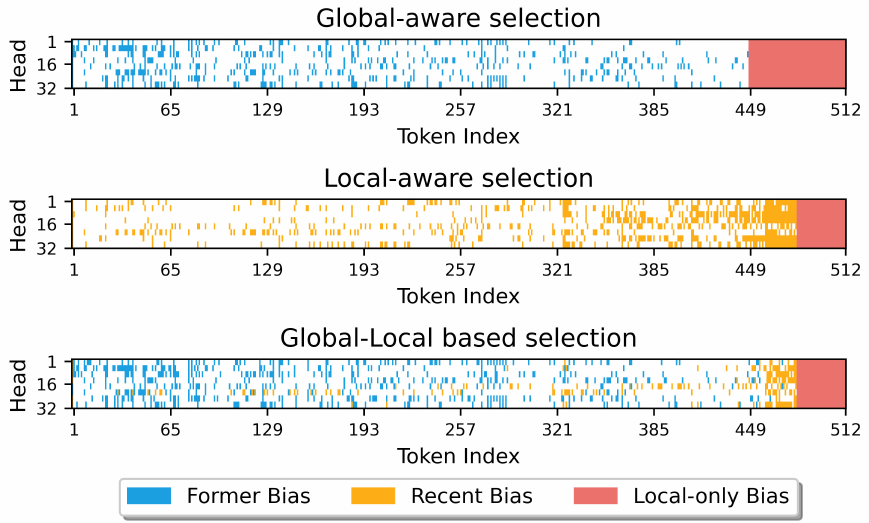}
        \caption{Layer 31 on \textit{hotpotqa} using Llama3.}
        \label{selection_hotpotqa_llama3_2}
    \end{subfigure}
    
    \caption{Token selection patterns visualization on different layers, datasets and models. Global-aware selection tends to select former tokens while local-aware selection prefers recent tokens. Global-Local based selection can balance these two approaches.}
    \label{fig:selection_pattern}
\end{figure*}

To further analysis the how tokens are selected according to Global-Local score, we draw the proportion of selected tokens derived from global-aware, local-aware and local-only selection in Figure \ref{fig:selection_distribution}. The token numbers are averaged on head dimension. If selecting Top-256 from 4096 tokens, the number of tokens sourced from global-aware and local-aware sources is roughly 1:1. If selecting Top-1024 from 4096 tokens, this ratio is approximately 7:3. This implies that when merging 1024 tokens into 256 tokens, more global-aware tokens are merged. Since local tokens are changing during the decoding stage, our dynamic class centers can be indispensable.
\newpage

\end{document}